\documentclass{ieeeaccess}
\usepackage{cite}
\usepackage{amsmath,amssymb,amsfonts}
\usepackage{algorithmic}
\usepackage{graphicx, array, multirow}
\usepackage{textcomp}
\usepackage{bm}
\usepackage{booktabs}
\usepackage{pifont}
\newcommand{\etal}{\textit{et al.}}
\newcommand{\xmark}{\ding{55}}%
\newcommand{\cmark}{\ding{51}}%

\makeatletter
\AtBeginDocument{\DeclareMathVersion{bold}
\SetSymbolFont{operators}{bold}{T1}{times}{b}{n}
\SetSymbolFont{NewLetters}{bold}{T1}{times}{b}{it}
\SetMathAlphabet{\mathrm}{bold}{T1}{times}{b}{n}
\SetMathAlphabet{\mathit}{bold}{T1}{times}{b}{it}
\SetMathAlphabet{\mathbf}{bold}{T1}{times}{b}{n}
\SetMathAlphabet{\mathtt}{bold}{OT1}{pcr}{b}{n}
\SetSymbolFont{symbols}{bold}{OMS}{cmsy}{b}{n}
\renewcommand\boldmath{\@nomath\boldmath\mathversion{bold}}}
\makeatother

\def\BibTeX{{\rm B\kern-.05em{\sc i\kern-.025em b}\kern-.08em
    T\kern-.1667em\lower.7ex\hbox{E}\kern-.125emX}}

\begin{document}
\history{Date of publication xxxx 00, 0000, date of current version xxxx 00, 0000.}
\doi{10.1109/ACCESS.2024.0429000}

\title{SoGAR: Self-supervised Spatiotemporal Attention-based Social Group Activity Recognition}
\author{\uppercase{Naga VS Raviteja Chappa}\authorrefmark{1}, \IEEEmembership{Member, IEEE},
{Pha Nguyen}\authorrefmark{1}, {Alexander H Nelson}\authorrefmark{1}, {Han-Seok Seo}\authorrefmark{2}, {Xin Li}\authorrefmark{4}, {Page D Dobbs}\authorrefmark{3} and Khoa Luu\authorrefmark{1},
\IEEEmembership{Member, IEEE}}

\address[1]{Department of EECS, University of Arkansas, Fayetteville, USA (e-mail: nchappa@uark.edu, panguyen@uark.edu, ahnelson@uark.edu, khoaluu@uark.edu)}
\address[2]{Department of Food Science, University of Arkansas, Fayetteville, USA (e-mail: hanseok@uark.edu)}
\address[3]{Department of Computer Science, University at Albany, Albany, NY USA (e-mail: xli48@albany.edu)}
\address[4]{Department of Health, Human Performance and Recreation, University of Arkansas, Fayetteville, USA (e-mail: pdobbs@uark.edu)}

\tfootnote{This paragraph of the first footnote will contain support
information, including sponsor and financial support acknowledgment. For
example, ``This work was supported in part by the U.S. Department of
Commerce under Grant BS123456.''}

\markboth
{Author \headeretal: Preparation of Papers for IEEE TRANSACTIONS and JOURNALS}
{Author \headeretal: Preparation of Papers for IEEE TRANSACTIONS and JOURNALS}

\corresp{Corresponding author: Naga VS Raviteja Chappa (e-mail: nchappa@uark.edu).}

\begin{abstract}
This paper introduces a novel approach to Social Group Activity Recognition (SoGAR) using Self-supervised Transformers network that can effectively utilize unlabeled video data. To extract spatio-temporal information, we create local and global views with varying frame rates. Our self-supervised objective ensures that features extracted from contrasting views of the same video are consistent across spatio-temporal domains. Our proposed approach efficiently uses transformer-based encoders to alleviate the weakly supervised setting of group activity recognition. By leveraging the benefits of transformer models, our approach can model long-term relationships along spatio-temporal dimensions. Our proposed SoGAR method achieves state-of-the-art results on three group activity recognition benchmarks, namely JRDB-PAR, NBA, and Volleyball datasets, surpassing the current state-of-the-art in terms of F1-score, MCA, and MPCA metrics.
\end{abstract}


\titlepgskip=-21pt

\maketitle

\section{Introduction}
\label{sec:introduction}
Group activity recognition (GAR) has emerged as an important problem in computer vision, with numerous applications in sports video analysis, video monitoring, and social scene understanding. Unlike conventional action recognition methods that focus on identifying individual actions, GAR aims to classify the actions of a group of people in a given video clip as a whole. This requires a deeper understanding of the interactions between multiple actors, including accurate localization of actors and modeling their spatiotemporal relationships~\cite{wang2016temporal, carreira2017quo, wang2018non, Ranasinghe_2022_CVPR}. As a result, GAR poses fundamental challenges that need to be addressed in order to develop effective solutions for this problem. In this context, the development of novel techniques for group activity recognition has become an active area of research in computer vision.

\begin{figure}[!t]
    \centering
\includegraphics[width=0.47\textwidth]{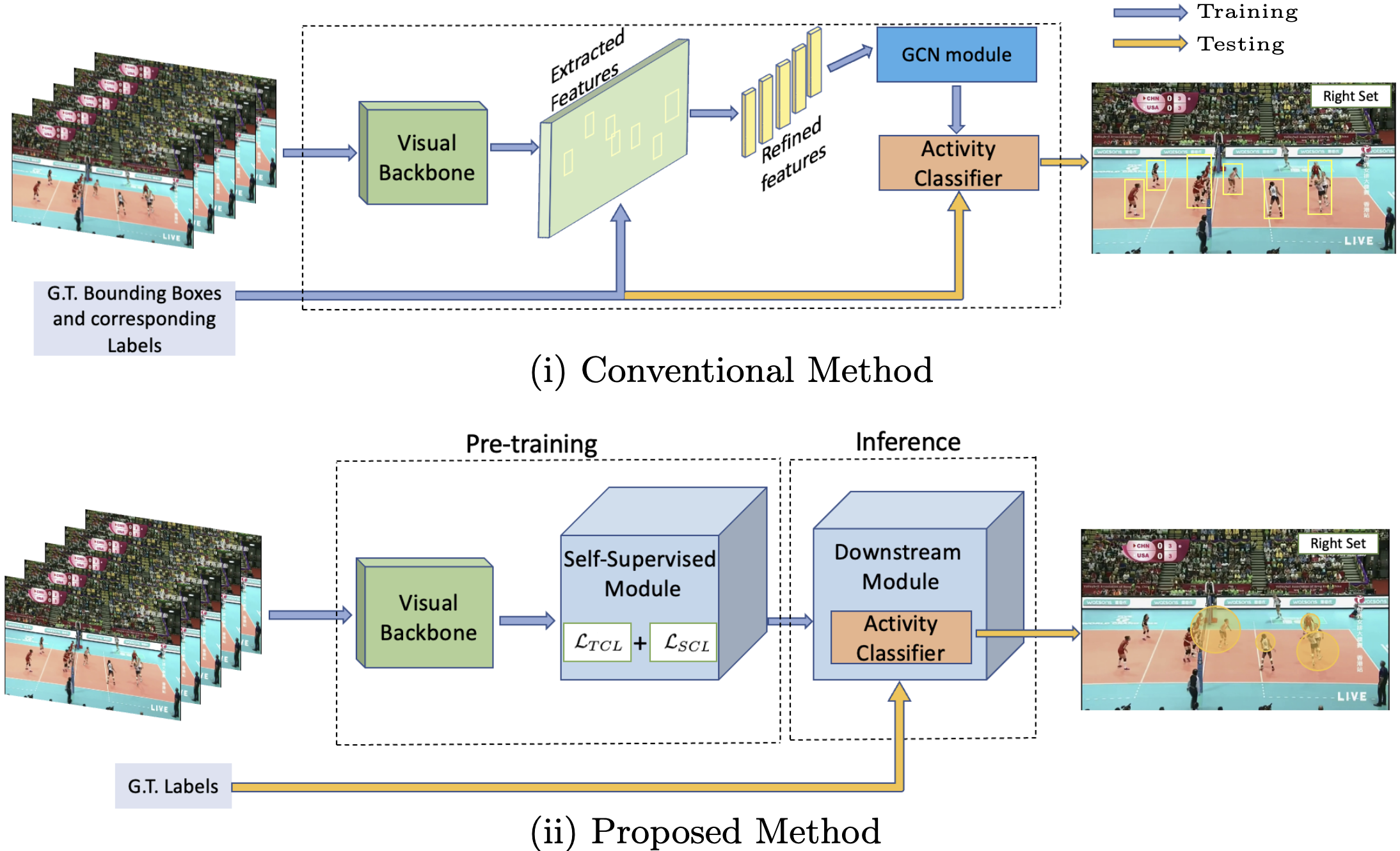}
    \caption{{Overview of conventional and proposed methods for social activity recognition. The labels in the right image show the predicted labels.}}
    \label{fig:page1}
\end{figure}

Existing methods for GAR require ground-truth bounding boxes and action class labels for training and testing~\cite{ibrahim2016hierarchical, wu2019learning, hu2020progressive, gavrilyuk2020actor, pramono2020empowering, ehsanpour2020joint, yan2020higcin, yuan2021learning, li2021groupformer}. Bounding box labels are used to extract actor features and their spatio-temporal relations, which are then aggregated to form a group-level video representation for classification. However, the reliance on bounding boxes and substantial data labeling annotations severely limit their applications.

To address these limitations, some methods simultaneously train person detection and group activity recognition using bounding box labels~\cite{bagautdinov2017social, zhang2019fast}. Another approach is weakly supervised GAR (WSGAR) learning~\cite{yan2020social, kim2022detector}, which does not require individual actor-level labels for training and inference.

Yan~\etal~\cite{yan2020social} proposed WSGAR learning approach that uses a pre-trained detector to generate actor box suggestions and learn to eliminate irrelevant possibilities. However, this method suffers from missing detections when actors are occluded. Kim~\etal~\cite{kim2022detector} introduced a detector-free method that captures actor information using partial contexts of token embeddings, but this method can only learn when there is movement in consecutive frames.
Moreover, Kim~\etal~\cite{kim2022detector} did not consider the consistency of temporal information among different tokens. Hence, there is a need for a GAR approach that can capture temporal information accurately without the limitations of bounding box annotations or detector-based methods.

\noindent\textbf{Contributions of this Work:}
In this paper, we propose a new approach to Social Group Activity Recognition called (SoGAR). Our method is unique in that it does not require ground-truth labels during pre-training, and it doesn't rely on an object detector. Instead, our approach uses motion as a supervisory signal from the RGB data modality. Our approach is able to effectively reduce the extensive supervision present in the conventional methods, as demonstrated in Fig.~\ref{fig:page1}. In fact, our method outperforms the DFWSGAR approach introduced by Kim et al. \cite{kim2022detector}. We also present the comparison of different properties between our approach and other previous methods in Table~\ref{tab:summary}.
To handle varying spatial and temporal details within the same deep network, we use a video transformer-based approach, as described in \cite{gberta_2021_ICML}. This approach allows us to take advantage of varying temporal resolutions within the same architecture. Additionally, the self-attention mechanism in video transformers can capture local and global long-range dependencies in both space and time, providing much larger receptive fields compared to standard convolutional kernels \cite{naseer2021intriguing}.

The proposed SoGAR method differs from the previous methods by leveraging the correspondences from spatio-temporal features which enables the learning of long-range dependencies in both space and time domains. To facilitate this, we introduce a novel self-supervised learning strategy that does temporal collaborative learning and spatiotemporal cooperative learning. This is achieved through the proposed loss functions mentioned in \ref{subsec:loss}, which match the global features from the whole video sequence to the local features that are sampled in the latent space. Additionally, we utilize the bounding box information to localize the attention of the framework for better learning to improve overall performance. Our proposed method achieves State-of-the-Art (SOTA) performance results on the JRDB-PAR~\cite{han2022panoramic}, NBA~\cite{yan2020social} and Volleyball~\cite{ibrahim2016hierarchical} datasets using only the RGB inputs. We conducted extensive experiments and will publish the code for our method.

\section{Related Work}

\subsection{Group Activity Recognition (GAR)}
In the field of action recognition, group action recognition has become an increasingly popular topic of research due to its wide range of applications in various fields, such as video surveillance, human-robot interaction, and sports analysis. GAR aims to identify the actions performed by a group of individuals and the interactions between them.

Initially, researchers in the field of GAR used probabilistic graphical methods and AND-OR grammar methods to process the extracted features~\cite{amer2012cost,amer2013monte,amer2014hirf,amer2015sum,lan2011discriminative,lan2012social,shu2015joint,wang2013bilinear}. However, with the advancement of deep learning techniques, methods involving convolutional neural networks (CNN) and recurrent neural networks (RNN) achieved outstanding performance due to their ability to learn high-level information and temporal context~\cite{bagautdinov2017social,deng2016structure,ibrahim2016hierarchical,ibrahim2018hierarchical,li2017sbgar,qi2018stagnet,shu2019hierarchical,wang2017recurrent,yan2018participation}.

Recent methods for identifying group actions typically utilize attention-based models and require explicit character representations to model spatial-temporal relations in group activities~\cite{ehsanpour2020joint,gavrilyuk2020actor,hu2020progressive,li2021groupformer,pramono2020empowering,wu2019learning,yan2020social,yuan2021spatio, chappa2024hatt, chappa2024ligar, chappa2024react, chappa2024flaash, truong2022otadapt}. For example, graph convolution networks are used to learn spatial and temporal information of actors by constructing relational graphs, and spatial and temporal relation graphs are used to infer actor links. Clustered attention is used to capture contextual spatial-temporal information, and transformer encoder-based techniques with different backbone networks are used to extract features for learning actor interactions from multimodal inputs~\cite{gavrilyuk2020actor}. Additionally, MAC-Loss~\cite{han2022dual}, a combination of spatial and temporal transformers in two complimentary orders, has been proposed to enhance the learning effectiveness of actor interactions and preserve actor consistency at the frame and video levels. Tamura~\etal~\cite{tamura2022hunting} introduces a framework without using heuristic features for recognizing social group activities and identifying group members. This information is embedded into the features, allowing for easy identification.

Overall, these recent advancements in GAR have made significant progress toward recognizing complex actions performed by a group of individuals in various settings. 
\begin{table*}[t]
\centering
\caption{ \textbf{Comparisons in the properties between our proposed approach and other methods}. Actor Relation Learning (ARL), Convolutional Neural Networks (CNN), Graph Neural Networks (GNN), Graph Convolutional Networks (GCN), Transformer (TF), TimeSformer (TSformer), Vision Transformer (ViT), Space \& Time (ST), Group Activity (G.A.), Individual Actions (I.A.), Bounding Boxes (B.B.)}
\small
\resizebox{\textwidth}{!}{\begin{tabular}{c|c|c|c|c}
\hline
\textbf{Methods} & \textbf{Architecture}              & \textbf{Source Label} & \textbf{Learning Mechanism} & \textbf{ARL Module} \\ \hline
ARG \cite{wu2019learning}        & CNN + GCN                          & G.A., I.A., B.B.  & Fully Supervised & Graph Relational Reasoning                                                                  \\ \hline
HiGCIN \cite{yan2020higcin}     & CNN + GNN                          & G.A., I.A., B.B. & Fully Supervised & Graph Relational Reasoning                                                                     \\ \hline
AT \cite{gavrilyuk2020actor}             & CNN + TF                              & G.A., I.A., B.B. &           Fully Supervised                    &       Joint ST Attention                        \\ \hline
DIN \cite{yuan2021spatio}          & CNN + GNN                          & G.A., I.A., B.B. & Fully Supervised & Graph Relational Reasoning                                 \\ \hline
GroupFormer \cite{li2021groupformer}         & CNN + TF                          & G.A., I.A., B.B.  & Fully Supervised & Clustering     \\ \hline
Dual-AI \cite{han2022dual}         & CNN + TF                          & G.A., I.A., B.B.  & Fully Supervised & Joint ST Attention         \\ \hline
SAM \cite{yan2020social}         & CNN + GCN                          & G.A., B.B.  & Weakly Supervised &  Graph Relational Reasoning   \\ \hline
DFWSGAR \cite{kim2022detector}         & CNN + TF                          & G.A. (Training \& Testing) & Weakly Supervised & Joint ST Attention   \\ \hline
\textbf{Ours}  & \textbf{ViT + TSformer} & \textbf{G.A.(Testing)}  & \textbf{Self-Supervised} & \textbf{Divided ST Attention}      \\ \hline
\end{tabular}}
\label{tab:summary}
\vspace{0mm}
\end{table*}

\begin{figure}[!t]
    \centering
\includegraphics[width=0.47\textwidth]{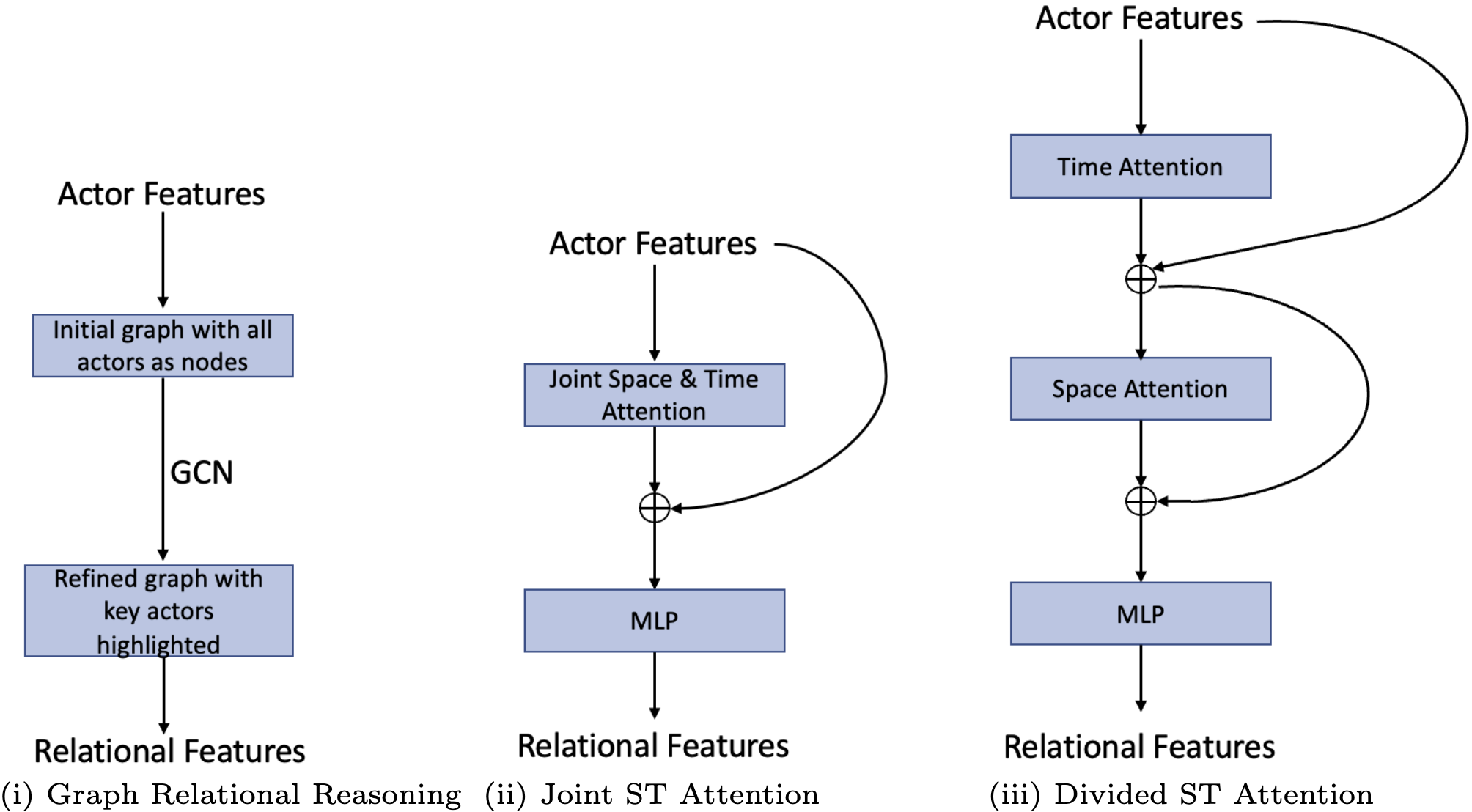}
    \caption{{Comparison of Actor Relational Learning (ARL) Modules}}
     \vspace{-0.2in}
    \label{fig:ARL}
\end{figure}

\noindent\textbf{Weakly supervised group activity recognition (WSGAR).} Various techniques have been developed to address the problem of WSGAR with limited supervision, like training detectors within the framework using bounding boxes. WSGAR is one approach that does not rely on bounding box annotations during training or inference and includes an off-the-shelf item detector in the model. Traditional GAR approaches require accurate annotations of individual actors and their actions, which can be challenging and time-consuming to obtain. Weakly supervised methods aim to relax these requirements by learning from more readily available data such as activity labels, bounding boxes, or even video-level labels. Zhang et al.~\cite{zhang2021multi} proposed a technique that employs activity-specific characteristics to enhance WSGAR. It is not particularly designed for GAR. Kim et al.~\cite{kim2022detector} proposed a detector-free approach that uses transformer encoders to extract motion features. We propose a self-supervised training method specialized for WSGAR and does not necessitate actor-level annotations, object detectors, or labels.

\noindent\textbf{Transformers in Vision}.
The transformer architecture was first introduced by Vaswani~\etal~\cite{vaswani2017attention} for sequence-to-sequence machine translation, and since then, it has been widely applied to various natural language processing tasks. Dosovitskiy~\etal~\cite{dosovitskiy2020image} introduced a transformer architecture not based on convolution for image recognition tasks. Several works~\cite{li2021ffa, yuan2021tokens,liu2021swin,wang2021pyramid} used transformer architecture as a general backbone for various downstream computer vision tasks, achieving remarkable performance progress. In the video domain, many approaches~\cite{han2020mining, arnab2021vivit, li2022uniformer, bertasius2021space,fan2021multiscale,patrick2021keeping} utilize spatial and temporal self-attention to learn video representations effectively. Bertasius~\etal~\cite{bertasius2021space} explored different mechanisms of space and time attention to learn spatiotemporal features efficiently. Fan et al. \cite{fan2021multiscale} used multiscale feature aggregation to improve the learning performance of features. Patrick~\etal~\cite{patrick2021keeping} introduced a self-attention block that focuses on the trajectory, which tracks the patches of space and time in a video transformer.

\section{The Proposed Method}
\begin{figure*}
    \centering
    \includegraphics[width=1.0\textwidth]{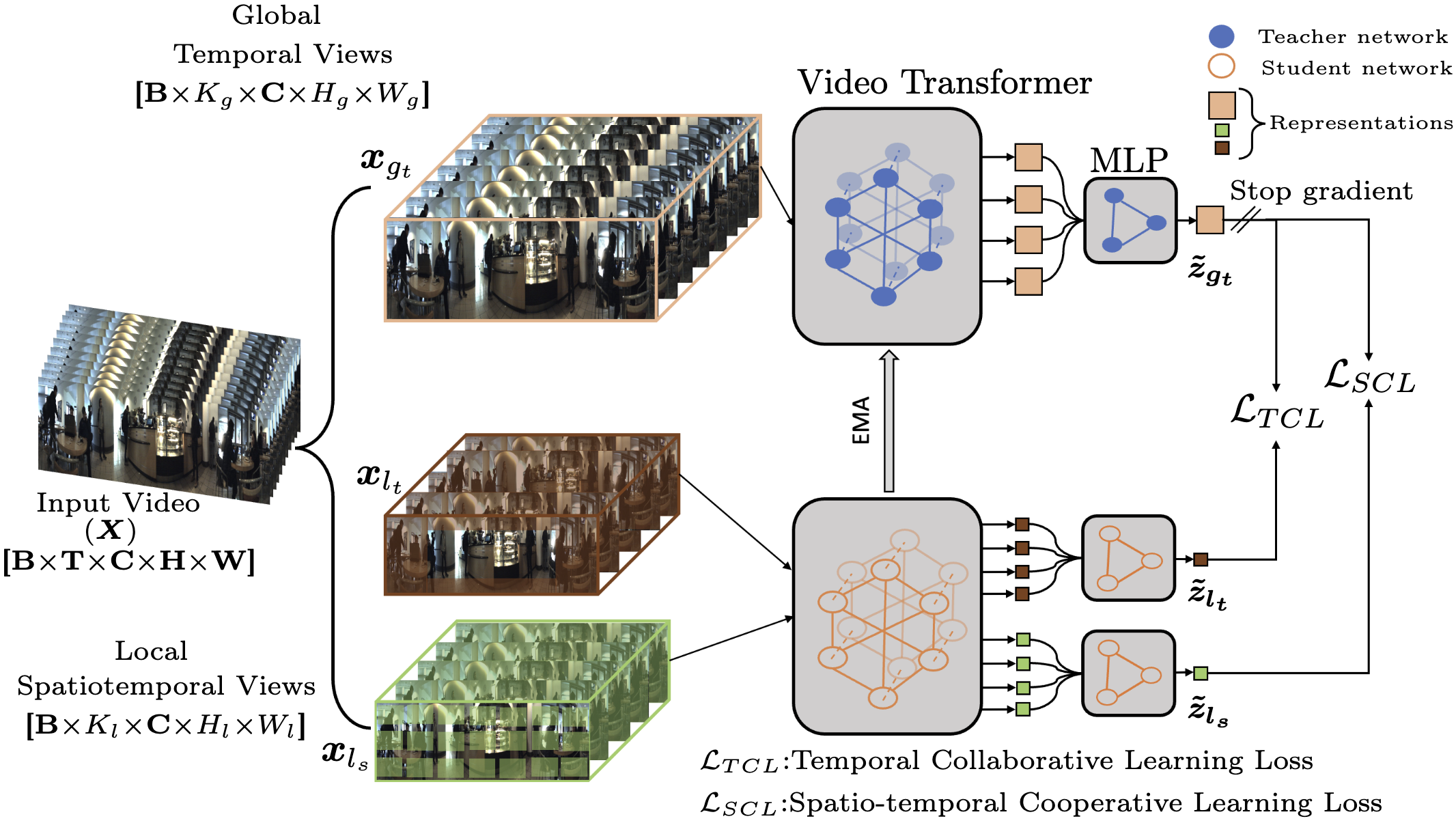}
    \caption{\textbf{The proposed SoGAR framework} adopts a sampling strategy that divides the input video into global and local views in temporal and spatial domains. Since the video clips are sampled at different rates, the global and local views have distinct spatial characteristics and limited fields of view and are subject to spatial augmentations. The teacher network takes in global views ($\bm{x}_{{g}{t}}$) to generate a target, while the student network processes local views ($\bm{x}_{{l}{t}}$ \& $\bm{x}_{{l}{s}}$), where $K{l}$ $\le$ $K_{g}$. We update the network weights by matching the student local views to the target teacher global views, which involves both \emph{Temporal Collaborative Learning} and \emph{Spatio-temporal Cooperative Learning}. To accomplish this, we employ a standard ViT-Base backbone with separate space-time attention \cite{gberta_2021_ICML} and an MLP that predicts target features from student features.}
    \label{fig:framework}
\end{figure*}

The framework presented in this paper aims to recognize social group activities in a video without depending on a detector or person-bounding boxes. The proposed method follows a self-supervised training approach within the teacher-student framework for social group activity recognition, as depicted in Fig. \ref{fig:framework}.

Our method for video representation learning for social group activity recognition differs from other contrastive learning approaches by processing two clips from the same video while altering their spatial-temporal characteristics without requiring memory banks. This approach allows us to capture the intricate and ever-changing nature of group activities where multiple individuals may be moving in different directions and performing different actions simultaneously.

To train our model, we propose a novel loss formulation that matches the features of two distinct clips, thereby enforcing consistency in spatial and temporal changes within the same video. Our loss function encourages the model to learn robust representations that can handle variations in spatial and temporal contexts.

The proposed {SoGAR} framework is described in detail in the following sections. We demonstrate the effectiveness of our method on the newly proposed JRDB-PAR dataset~\cite{han2022panoramic} along with NBA~\cite{yan2020social},  and Volleyball~\cite{ibrahim2018hierarchical} datasets.
 
\subsection{Self-Supervised Training}\label{subsec:ssltraining}

Videos of social group activities capture rich temporal and spatial information, which is essential for accurate recognition. However, this high temporal dimensionality also makes it challenging to capture the various motion and spatial characteristics of group activities, such as 2p.-fail. (from NBA dataset~\cite{yan2020social}) or l-winpoint (from Volleyball dataset~\cite{ibrahim2016hierarchical}). To address this challenge, we propose a novel approach that involves predicting different video clips with varying temporal characteristics from each other in the feature space. This approach allows us to learn contextual information that defines the underlying distribution of videos, making the network invariant to motion, scale, and viewpoint variations.

Our self-supervised training framework for video representation learning is formulated as a motion prediction problem consisting of three key components. First, we generate multiple temporal views with different numbers of clips with varying motion characteristics from the same video. Second, we vary the spatial characteristics of these views by generating local and global spatial fields of the sampled clips. Finally, we introduce a loss function that matches the varying views across spatial and temporal dimensions in the latent space.

The proposed approach for social group activity recognition involves predicting multiple video clips with varying temporal and spatial characteristics from a single video. This is achieved through a self-supervised motion prediction problem with three key components: generating multiple temporal views with different numbers of clips and varying motion characteristics, varying the spatial characteristics of these views by generating local and global spatial fields of the sampled clips, and introducing a loss function that matches the varying views across spatial and temporal dimensions in the latent space. By learning contextual information and making accurate predictions even in the presence of various motion, scale, and viewpoint variations, the network becomes invariant to these variations and can capture the complex and dynamic nature of social group activities.

\begin{figure}[!t]
    \centering
\includegraphics[width=0.3\textwidth]{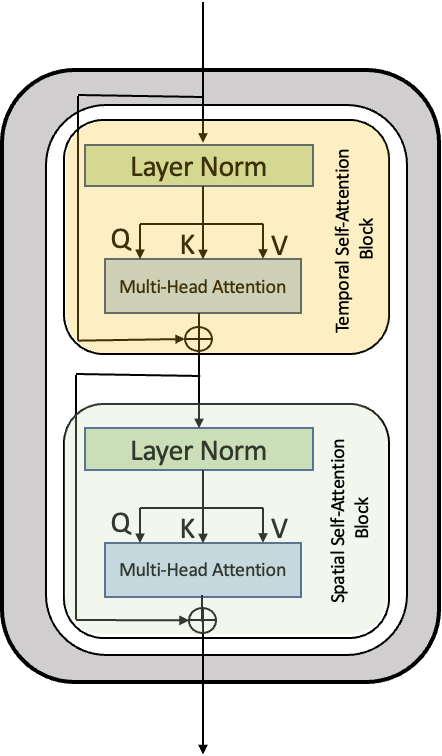}
    \caption{{Video Transformer Block}}
     \vspace{-0.2in}
    \label{fig:vtn}
\end{figure}

\subsubsection{Prediction of motion via Self-Supervised Learning}
\label{subsec:mo_pred}
The temporal dimension of a video is a crucial factor that can significantly affect the motion context and perception of actions captured in the content. For example, the frame rate can capture subtle nuances of body movements and affect the perception of actions, such as walking slowly versus walking quickly. Traditionally, video clips are sampled at a fixed frame rate, which may not be suitable for capturing different motion characteristics of the same action.

Our proposed approach introduces the concept of "temporal views," which refers to a collection of clips sampled at a specific video frame rate. By generating different views with varying resolutions, we can capture different motion characteristics of the same action and learn contextual information about motion from a low frame rate input. To create motion differences among these views, we randomly sample them and process them using our ViT models. The number of temporal tokens ($T$) input to ViT varies in different views, allowing us to handle variability in temporal resolutions with a single ViT model.

In addition to varying temporal resolution, we vary the resolution of clips across the spatial dimension within these views. This means that the spatial size of a clip can be lower than the maximum spatial size (224), which can also decrease the number of spatial tokens. Using vanilla positional encoding~\cite{vaswani2017attention}, our approach can handle such variability in temporal resolutions with a single ViT model, unlike similar sampling strategies used under multi-network settings~\cite{feichtenhofer2019slowfast, kahatapitiya2021coarse}. 

\subsubsection{Establishing Correspondences Across Different Views}
\label{subsec:cross_view_corrspondences}
Our proposed training strategy seeks to establish the interrelation between a given video's temporal and spatial dimensions. To achieve this, we introduce novel cross-view correspondences by manipulating the field of view during the sampling process. In particular, we generate global and local temporal views from a given video clip to facilitate learning these correspondences.

The global temporal views ($\bm{x}_{{g}_{t}}$) are generated by randomly sampling $K_{g}$ frames from a video clip with a fixed spatial size of $W_{global}$ and $H_{global}$. These views are then fed into the teacher network, which produces an output represented by $\bm{\Tilde{z}_{g_{t}}}$.

On the other hand, the local spatiotemporal views ($\bm{x}_{{l}_{t}}$ and $\bm{x}_{{l}_{s}}$) cover a limited portion of the video clip along both spatial and temporal dimensions. We generate these local temporal views by randomly selecting several frames ($K_{l}$), which is less than or equal to the number of frames in the global temporal views ($K_{g}$), with a spatial size fixed to $W_{local}$ and $H_{local}$. These views are then fed into the student network, which produces two outputs denoted by $\bm{\Tilde{z}_{l_{t}}}$ and $\bm{\Tilde{z}_{l_{s}}}$, respectively.

We apply various data augmentation techniques to the spatial dimension by applying color jittering and gray scaling with probability 0.8 and 0.2, respectively, to all temporal views. Moreover, we apply Gaussian blur and solarization with probability 0.1 and 0.2, respectively, to global temporal views.

Our approach is based on the idea that training the model to predict a global temporal view of a video from a local temporal view in the latent space can help the model capture high-level contextual information. More specifically, our method encourages the model to consider both the spatial and temporal context of the video, where the spatial context denotes the possibilities surrounding a given spatial crop, and the temporal context denotes possible previous or future clips from a given temporal crop. It is essential to note that spatial correspondences also involve a temporal component, as our approach seeks to predict a global view at timestamp $t=j$ from a local view at timestamp $t=i$. To enforce these cross-view correspondences, we use a similarity objective that predicts different views from each other.

\subsection{The Proposed Objective Function}\label{subsec:loss}
Our model aims to predict different views of the same video, capturing various spatial-temporal variations. To achieve this, we train our model with an objective function that leverages global and local temporal and spatial views.

Let $\bm{X}={\bm{x}_t}^T$ be a video consisting of $T$ frames, where $\bm{x}_{{g}_{t}}$, $\bm{x}_{{l}_{t}}$, and $\bm{x}{{l}_{s}}$ represent global temporal views, local temporal views, and local spatial views, respectively. Specifically, $\bm{x}_{{g}_{t}}$ contains $K{g}$ frames, while $\bm{x}_{{l}_{t}}$ and $\bm{x}_{{l}_{s}}$ both contain $K_{l}$ frames, where $K_{l} \le K_{g}$ and $K_{g}$ and $K_{l}$ are the numbers of frames for teacher and student (global and local) inputs. We randomly sample $K_{g}$ global and $K_{l}$ local temporal views as described in \ref{subsec:cross_view_corrspondences}.
The student and teacher models process the temporal views to obtain class tokens or features $\bm{z}_g$ and $\bm{z}_l$. We then normalize these class tokens to facilitate training with the objective function.

\begin{equation}
     \bm{\Tilde{z}}^{(i)} = \frac{\text{exp}(\bm{z}^{(i)}) / \tau}{\sum_{i=1}^n \text{exp}(\bm{z}^{(i)})/ \tau },
\end{equation}
where $\tau$ is a temperature parameter used to control the sharpness of the exponential function \cite{caron2021emerging} and $\bm{z}^{(i)}$ is each element in $\bm{\Tilde{z}^{(i)}}\in\mathbb{R}^{n}$.  

\textbf{Temporal Collaborative Learning Loss (TCL):}
Our $\bm{x}_{{g}_{t}}$ have the same spatial size but differ in temporal content because the number of clips/frames is randomly sampled for each view. One of the $\bm{x}_{{g}_{t}}$ always passes through the teacher model that serves as the target label. We map the student's $\bm{x}_{{l}_{t}}$ with the teacher's $\bm{x}_{{g}_{t}}$ to create a global-to-local temporal loss as in Eqn. \eqref{eq:global_to_global_loss}.
\begin{align}\label{eq:global_to_global_loss}
    \mathcal{L}_{TCL} &=  -\bm{sg(\Tilde{z}}_{g_{t}}) * log(\bm{\Tilde{z}}_{l_{t}}),
\end{align}
where $\bm{\Tilde{z}_{g_{t}}}$ and $\bm{\Tilde{z}_{l_{t}}}$ are the tokens of the class for $\bm{x}_{{g}_{t}}$ and $\bm{x}_{{l}_{t}}$ produced by the teacher and student, and $sg$ is the stochastic gradient respectively.

\textbf{Spatio-temporal Cooperative Learning Loss (SCL):} 
The local temporal views $\bm{x}_{{l}_{t}}$ in our approach have a smaller field of vision compared to the global temporal views $\bm{x}_{{g}_{t}}$, both along the spatial and temporal dimensions. Despite this, the number of local views is four times higher than that of global views. The student model processes all the local views $\bm{x}_{{l}_{s}}$, while the teacher model processes only the global views $\bm{x}_{{g}_{t}}$, which serve as the target. To create the loss function, the local views are mapped to the global views using the teacher model, as described in \ref{eq:local_to_global_loss}.

\begin{align}\label{eq:local_to_global_loss}
    \mathcal{L}_{SCL} &= \sum_{n=1}^{q} -\bm{sg(\Tilde{z}}_{g_{t}}) * log(\bm{\Tilde{z}}^{(n)}_{l_{s}}),
\end{align}
where $\bm{\Tilde{z}_{l_{s}}}$ are the tokens of the class for $\bm{x}_{{l}_{s}}$ produced by the student and $q$ represents the number of local temporal views set to sixteen in all our experiments. The overall loss to train our model is simply a linear combination of both losses, as in Eqn. \eqref{eq:global_to_global_loss} and Eqn. \eqref{eq:local_to_global_loss}, given as in Eqn. \eqref{eq:totalloss}.
\begin{equation} \label{eq:totalloss}
    \mathcal{L} =  \mathcal{L}_{TCL} +  \mathcal{L}_{SCL}
\end{equation}
\subsection{Inference}\label{subsec:Inference}
\begin{figure}[!t]
    \centering
    \includegraphics[width=0.47\textwidth]{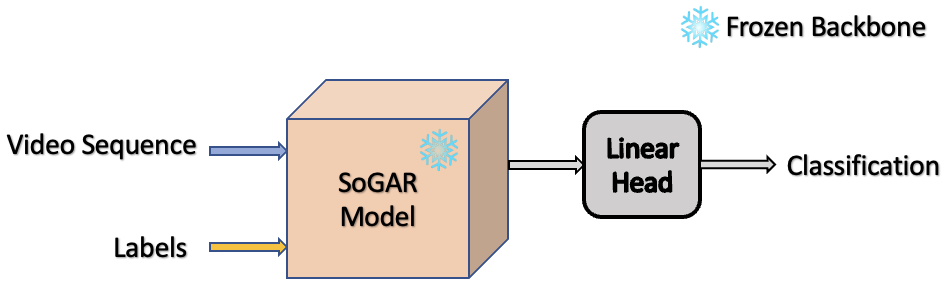}
\caption{\textbf{Inference}. We input the video sequence along with their corresponding labels. The output from the model is fed to the downstream task classifier.}
    \vspace{-0.1in}
    \label{fig:my_label1}
\end{figure}

Our inference framework is depicted in Fig. \ref{fig:my_label1}. In this stage, we perform fine-tuning of the self-supervised model that was trained earlier. Specifically, we utilize the pre-trained SoGAR model and fine-tune it with the available labels. This is followed by a linear classifier, and the resulting model is applied to downstream tasks to enhance the overall performance.

\section{Experiments}

\subsection{Datasets}
\noindent\textbf{Volleyball Dataset}\cite{ibrahim2016hierarchical} is composed of 55 videos, containing a total of 4,830 labeled clips, including 3,493 for training and 1,337 for testing. The dataset provides annotations for both individual actions and group activities with corresponding bounding boxes. However, in our WSGAR experiments, we only focus on the group activity labels and exclude the individual action annotations. To evaluate our model, we use Multi-class Classification Accuracy (MCA) and Merged MCA metrics. The Merged MCA metric merges the right set and right pass classes into the right pass-set and the left set and left pass classes into the left pass-set, as in previous works like SAM~\cite{yan2020social} and DFWSGAR~\cite{kim2022detector}, to ensure a fair comparison with existing methods.

\noindent\textbf{NBA Dataset}\cite{yan2020social} used in our experiments contains a total of 9,172 labeled clips from 181 NBA videos, where 7,624 clips are for training and 1,548 for testing. The dataset only provides annotations for group activities and lacks information about individual actions or bounding boxes. For evaluating the model, we use the Multi-class Classification Accuracy (MCA) and Mean Per Class Accuracy (MPCA) metrics. The MPCA metric is used to address the issue of class imbalance in the dataset.

\noindent\textbf{JRDB-PAR Dataset}\cite{han2022panoramic} containing 27 categories of individual actions such as walking, talking, etc., 11 categories of social group activities, and 7 categories of global activities. The dataset consists of 27 videos, which are split into 20 for training and 7 for testing, following the training/validation splitting in JRDB dataset~\cite{ehsanpour2022jrdb}. In total, the dataset contains 27,920 frames with over 628k human bounding boxes. For annotation and evaluation, uniformly sampled keyframes (one keyframe in every 15 frames) are selected, which is consistent with other group activity datasets like CAD~\cite{choi2011learning} and Volleyball~\cite{ibrahim2018hierarchical}. The dataset uses multi-class labels for activity annotation, with each individual/group/frame having multiple activity labels. Following~\cite{han2022panoramic}, we use the precision, recall, and F1-score (denoted as $\mathcal{P}_g$, $\mathcal{R}_g$, $\mathcal{F}_g$) for evaluation, since social group activity recognition can be considered as a multi-label classification problem. 

\subsection{Deep Network Architecture}

Our video processing technique employs a Vision Transformer (ViT) \cite{gberta_2021_ICML} to apply attention to both the spatial and temporal dimensions of video clips. The ViT comprises 12 encoder blocks and can handle video clips of size $(B\times T\times C\times W\times H)$, where $B$ and $C$ denote the batch size and the number of color channels, respectively. The maximum spatial and temporal sizes are $W=H=480$ and $T=18$, respectively, indicating that we extract 18 frames from each video and resize them to $480\times 480$.
Our network architecture (see Fig.~\ref{fig:framework}) is designed to accommodate varying input resolution during training, including differences in frame rate, number of frames in a video clip, and spatial size. However, each ViT encoder block processes a maximum of 196 spatial and 16 temporal tokens, with each token having an embedding dimension of $\mathbb{R}^{m}$ \cite{dosovitskiy2020image}. In addition to these spatial and temporal input tokens, we include a single classification token within the architecture as a characteristic vector \cite{devlin2018bert}. This classification token captures the standard features learned by the ViT across the spatial and temporal dimensions of a given video. During training, we use varying spatial and temporal resolutions that satisfy $W \le 480$, $H \le 480$, and $T \le 18$, resulting in different spatial and temporal tokens. Finally, we apply a projection head to the class token of the last ViT encoder \cite{caron2021emerging, grill2020bootstrap}.
 
\textbf{Self-Distillation.} 
Our approach, depicted in Fig. \ref{fig:framework}, employs a teacher-student setup for self-distillation based on the methodology proposed in \cite{caron2021emerging, grill2020bootstrap}. The teacher and student models share the same architecture, consisting of a ViT backbone and a predictor MLP. However, only the student model is directly trained, while the teacher model is updated through an exponential moving average (EMA) of the student weights at each training step \cite{caron2021emerging}. This design allows us to use a unified network to process various input clips.

\subsection{Implementation Details}

To prepare the JRDB-PAR, NBA and Volleyball datasets for our analysis, we sampled frames at a rate of T ($K_{g}$) using segment-based sampling, as detailed in \cite{wang2016temporal}. Next, we resized the frames to $W_{g} = 480$ \& $H_{g} = 480$ for the teacher input and $W_{l} = 96$ \& $H_{l} =96$ for the student input. In the case of the Volleyball dataset, we set $K_{g}$ to 5 ($K_{l}\in{3,5}$), while for the NBA dataset, we set $K_{g}$ to 18 ($K_{l}\in{2,4,8,16,18}$). For JRD-PAR dataset, we used $K_{g}$ to 8 ($K_{l}\in{2,4,8,16,18}$). We initialized temporal attention weights randomly, while spatial attention weights were initialized using a ViT model trained self-supervised over ImageNet-1K \cite{imagenet}. This initialization scheme facilitated faster convergence of space-time ViT, as seen in the supervised setting \cite{gberta_2021_ICML}. We trained using an Adam optimizer \cite{kingma15adam} with a learning rate of $5\times10^{-4}$, scaled using a cosine schedule with a linear warm-up over five epochs \cite{Steiner2021HowTT, chen2021mocov3}. Additionally, we applied weight decay scaled from 0.04 to 0.1 during training. For the downstream task, we trained a linear classifier on our pretrained SPARTAN backbone. During training, the backbone was frozen, and we trained the classifier for 100 epochs with a batch size of 32 on a single NVIDIA-V100 GPU using SGD with an initial learning rate of 1e-3 and a cosine decay schedule. We also set the momentum to 0.9.

\subsection{Comparison with state-of-the-art methods}

\noindent\textbf{JRDB-PAR dataset}
We conducted a comparative study to evaluate our proposed approach alongside state-of-the-art methods in GAR and WSGAR using the JRDB-Par dataset. We involved fully supervised and weakly supervised settings to evaluate the dataset. The comparison results are presented in Table~\ref{tab:jrdpar}. In the fully supervised setting, our method outperforms the existing social group activity recognition frameworks significantly in all the metrics. In the weakly supervised setting, our proposed method outperformed existing GAR and WSGAR methods by a considerable margin, achieving 8.7 of $\mathcal{P}_g$, 12.7 of $\mathcal{R}_g$ and 9.9 of $\mathcal{F}_g$. Additionally, we evaluated this dataset using ResNet-18 and ViT-Base backbones, where ViT-Base proved to be better, which is analyzed in the ablation study section. Despite their impressive performance in WSGAR, our approach outperformed them all.
\begin{table}[ht]
		\caption{{Comparative results of the social group activity recognition on JRDB-PAR dataset~\cite{han2022panoramic}.}}
		\label{tab:jrdpar}  \vspace{10pt}
		
			\centering
			\renewcommand\tabcolsep{5pt}
			\footnotesize
			\begin{tabular}[*c]{l|ccc}
				\hline
				\multirow{2}{30pt}{Method}  &\multicolumn{3}{c}{{Group Activity}}  \\   \cline{2-4} 
				&$\mathcal{P}_g$   &$\mathcal{R}_g$  &$\mathcal{F}_g$ 
				 \\ \hline 
                \multicolumn{4}{c}{\textbf{Fully supervised}} \\
                \hline
				ARG~\cite{wu2019learning}  &34.6 &29.3 &30.7 \\
				SA-GAT~\cite{ehsanpour2020joint} 	&36.7 &29.9 &31.4 \\		
				JRDB-Base~\cite{ehsanpour2022jrdb} 	&44.6	&46.8	&45.1	\\			
				\textbf{Ours}   	&\textbf{49.3} 	&\textbf{47.1} 	&\textbf{48.7}	 \\
				\hline   
                \multicolumn{4}{c}{\textbf{Weakly supervised}} \\
                \hline 			
				AT\cite{gavrilyuk2020actor} 	&21.2 	&19.1 &19.8	 \\
				SACRF\cite{pramono2020empowering} 	&42.9 	&35.5  &37.6 \\
				Dynamic\cite{yuan2021spatio} 	&37.5 &27.1 &30.6	\\
				HiGCIN\cite{yan2020higcin} 	&39.3 	&30.1 &33.1	  \\		
				ARG\cite{wu2019learning} 			&26.9 	&21.5 & 23.3 \\
				SA-GAT\cite{ehsanpour2020joint}  	&28.6 	&24.0 	&25.5 \\	
				JRDB-Base\cite{ehsanpour2022jrdb} 	&38.4	&33.1	&34.8\\
				\textbf{Ours}  	&\textbf{47.1} 	&\textbf{45.8} 	&\textbf{44.9} \\
				\hline      
			\end{tabular} 

	\end{table}

\noindent\textbf{NBA dataset}
Table~\ref{table:SOTA_NBA} lists the outcomes of our comparison study on NBA dataset. Our approach outperforms existing GAR and WSGAR methods significantly, achieving 7.5\% MCA and 2.3\% MPCA. SAM's results~\cite{yan2020social} from~\cite{kim2022detector} are also listed. RGB frames are exclusively used as input to ensure a fair comparison across approaches and video backbones, including ResNet-18 TSM~\cite{lin2019tsm} and VideoSwin-T~\cite{liu2021video}. Comparing our approach to these strong backbones, our method prevails. Evaluating our proposed approach against current video backbones and state-of-the-art methods in GAR and WSGAR, our comparison study utilizes the NBA dataset. Notably, results of SAM~\cite{yan2020social} are referenced from~\cite{kim2022detector}.

\begin{table}[ht]
\caption{Comparisons with the State-of-the-Art GAR models and video backbones on the NBA dataset \cite{yan2020social}.
}
\begin{center}
\begin{tabular}{>{\arraybackslash}m{2.40cm} | >{\centering\arraybackslash}m{1.3cm}>{\centering\arraybackslash}m{0.85cm}>{\centering\arraybackslash}m{0.8cm}>{\centering\arraybackslash}m{0.9cm}}
\hline
Method                                                        & MCA       & MPCA      \\
\hline
\multicolumn{3}{c}{\textbf{Video backbone}} \\
\hline
TSM~\cite{lin2019tsm}                                        & 66.6      & 60.3      \\
VideoSwin~\cite{liu2021video}                                & 64.3      & 60.6      \\
\hline
\multicolumn{3}{c}{\textbf{GAR model}} \\
\hline
ARG~\cite{wu2019learning}                                      & 59.0      & 56.8      \\
AT~\cite{gavrilyuk2020actor}                                  & 47.1      & 41.5      \\
SACRF~\cite{pramono2020empowering}                            & 56.3      & 52.8      \\
DIN~\cite{yuan2021spatio}                                   & 61.6      & 56.0      \\

SAM~\cite{yan2020social}                          & 54.3      & 51.5      \\

DFWSGAR~\cite{kim2022detector}                                                          & 75.8     & 71.2     \\
\hline
\textbf{Ours}  &\textbf{83.3} & \textbf{73.5}\\
\hline
\end{tabular}
 
\end{center}
\vspace{-4mm}

\vspace{-1.5em}

\label{table:SOTA_NBA}
\vspace{3.5em}
\end{table}

\noindent
\textbf{Volleyball dataset.}
In the volleyball dataset, we reproduce results using only the RGB input and ResNet-18 backbone, respectively, to ensure a fair comparison. To have consistent comparison, we compare our approach against the latest GAR and WSGAR methods in two supervision levels: fully supervised and weakly supervised. The results show that our ResNet-18 trained model surpasses most fully supervised frameworks, showing a remarkable enhancement in MCA and MPCA metrics. The first and second sections display the outcomes of earlier techniques in fully supervised and weakly supervised contexts, respectively. Employing the ViT-Base backbone, our approach excels in weakly supervised conditions, outperforming all GAR and WSGAR models. By utilizing the transformer architecture to leverage spatiotemporal features, we achieve a significant lead of 2.4\% in MCA and 1.2\% in Merged MCA. Notably, these levels differ in their use of actor-level labels like ground-truth bounding boxes and individual action class labels during training and inference. In the weakly supervised setting, the group action classification labels are substituted with ground-truth bounding boxes of actors minus their corresponding actions. Table~\ref{table:SOTA_Volleyball} showcases the results. Additionally, our approach fares better than current GAR methods employing less comprehensive actor-level supervision, such as~\cite{bagautdinov2017social, yan2018participation, qi2018stagnet, gavrilyuk2020actor, pramono2020empowering}.

\begin{table}[ht]
\caption{Comparison with the state-of-the-art methods on the Volleyball dataset.~\cite{ibrahim2016hierarchical}}
\begin{center}
\begin{tabular}{>{\arraybackslash}m{2.4cm}| >  {\centering\arraybackslash}m{2.3cm} >{\centering\arraybackslash}m{0.9cm}>{\centering\arraybackslash}m{1.0cm}}

\hline
Method                             & Backbone              & MCA   & Merged MCA\\ [0.3ex]
\hline
\multicolumn{4}{c}{\textbf{Fully supervised}} \\
\hline
SSU~\cite{bagautdinov2017social}                 & Inception-v3          & 89.9  & - \\
PCTDM~\cite{yan2018participation}                & ResNet-18             & 90.3  & 94.3\\ %
StagNet~\cite{qi2018stagnet}                     & VGG-16                & 89.3  & - \\
ARG~\cite{wu2019learning}                        & ResNet-18             & 91.1 & \underline{95.1}\\ %
CRM~\cite{azar2019convolutional}        & I3D                   & 92.1  & - \\
HiGCIN~\cite{yan2020higcin}                      & ResNet-18             & 91.4  & - \\
AT~\cite{gavrilyuk2020actor}                     & ResNet-18             & 90.0 & 94.0\\ %
SACRF~\cite{pramono2020empowering}               & ResNet-18             & 90.7 & 92.7   \\ %
DIN~\cite{yuan2021spatio}                        & ResNet-18             & \underline{93.1}  & \textbf{95.6} \\ %
TCE+STBiP~\cite{yuan2021learning}        & VGG-16                & \textbf{94.1}  & - \\
GroupFormer~\cite{li2021groupformer}                      & Inception-v3             & \textbf{94.1}  & - \\

\hline
\multicolumn{4}{c}{\textbf{Weakly supervised}} \\
\hline 
PCTDM~\cite{yan2018participation}               & ResNet-18             & 80.5  & 90.0\\
ARG~\cite{wu2019learning}                       & ResNet-18             & 87.4  & 92.9\\
AT~\cite{gavrilyuk2020actor}                    & ResNet-18             & 84.3  & 89.6\\
SACRF~\cite{pramono2020empowering}              & ResNet-18             & 83.3  & 86.1  \\
DIN~\cite{yuan2021spatio}                       & ResNet-18             & 86.5  & 93.1\\
SAM~\cite{yan2020social}                        & ResNet-18             & 86.3  & 93.1\\

DFWSGAR~\cite{kim2022detector}                                         & ResNet-18             & 90.5  & 94.4\\
\hline
\multirow{2}{4em}{\textbf{Ours}} & ResNet-18 & 91.8 & 94.5 \\
 & ViT-Base & \textbf{93.1}& \textbf{95.9} \\
\hline
\end{tabular}

\end{center}


\label{table:SOTA_Volleyball}
\end{table}

\subsection{Ablation Study}
We conduct a thorough analysis of the various components that contribute to the effectiveness of our approach, which is an extension of analysis from~\cite{chappa2023spartan}. In particular, we assess the impact of five distinct elements:
a) Impact of different backbone networks,
b) Impact of knowledge distillation, and
c) Impact of ground-truth bounding box information

\noindent\textbf{Different Backbone Networks}: We investigated the effect of different backbone networks on our framework. We conducted the experiments presented in Table~\ref{tbl:ablation_backbone}. Our results show that ResNet-18 performs better than the other Convolutional Neural Network (CNN) backbones, but overall performance is optimal with ViT-Base backbone because the spatiotemporal features of the input video with varying views are well leveraged by the transformer architecture for videos~\cite{bertasius2021space}.  Also, when both networks share the same backbone, they perform better rather than having distinct backbone networks. 

\begin{table}[ht]
 \caption{\textbf{Different backbones.} The most optimal backbone for our framework is ViT-Base outperforming the other backbones.}
    \label{tbl:ablation_backbone}
\begin{center}
    
	\setlength{\tabcolsep}{7pt}
	\scalebox{1.0}[1.0]{
	\begin{tabular}{c|c|c|c}
		\toprule
		 
		\multirow{2}{*}{Backbone}  & JRDB-PAR & NBA & Volleyball \\
  &$(\mathcal{F}_g)$ & (MCA) & (MCA) \\\midrule
		Inception-v3   & 31.8   & 69.3   & 78.6   \\ 
		VGG-16   &  35.1  & 72.9   & 81.5   \\ 
		I3D   & 36.3   & {76.7}   & {85.8}   \\ 
        ResNet-18   & 39.6   & {78.1}   & {89.2} \\
        ViT-S  & {41.3}  & {80.2}   & {91.1} \\ \hline
        ViT-B   & \textbf{44.9}   & \textbf{83.3}   & \textbf{93.1}   \\ \bottomrule
	\end{tabular}
    
	}

\end{center}
\end{table}

\noindent\textbf{Impact of Knowledge Distillation (KD)}: To evaluate the effect of knowledge distillation, we conducted experiments as presented in Table~\ref{tbl:ablation_kd}. To be specific, we compared the performance of our approach in the absence of KD, i.e., the student and teacher networks learn independently, and there is no transfer of information from the student to teacher network. This shows very poor performance. Hence, KD is determined to be one of the key factors in the optimal performance of the proposed framework. This also proves that exponential moving average (EMA) aids feature learning across the networks to improve performance.

\begin{table}[ht]
\caption{\textbf{Impact of Knowledge Distillation (KD)}: The framework is proved to work better when there is knowledge distillation with EMA which infers student-teacher network learns the spatiotemporal features for different views on all the datasets.}
\begin{center}
\setlength{\tabcolsep}{5pt}
	\scalebox{1.0}[1.0]{
	\begin{tabular}{c|c|c|c}
		\toprule
		
		KD  & JRDB-PAR   & NBA    & Volleyball  \\  \midrule
		\xmark   &  34.2 & 75.2   & 86.4 \\
		\cmark   &  \textbf{44.9} & \textbf{83.3}   & \textbf{93.1} \\ \bottomrule
	\end{tabular}}
 
    \label{tbl:ablation_kd}
\end{center}
\end{table}

\noindent\textbf{Impact of ground-truth bounding box (G.T. BB's) information}: During the pre-training step, the social group activity recognition is highly leveraged by the actor localization information. So, we perform experiments as shown in Table~\ref{tbl:ablation_crops} to evaluate the performance of our method on this information. Specifically, we used random crops in the initial experiment in all the input views, which yields poor performance for JRDB-PAR and Volleyball datasets but the NBA dataset performs well as there is no bounding box information from the dataset. In contrast, we used the G.T. BB's exclusively without their corresponding labels for the other experiment to prove the optimal performance of our method.

\begin{table}[ht]
\caption{\textbf{Impact of ground-truth bounding box information (G.T. BB's))}: When we provided the bounding box information during the pre-training, it is proved that the performance is optimal rather than using random crops.}
\begin{center}

\setlength{\tabcolsep}{5pt}
	\scalebox{1.0}[1.0]{
	\begin{tabular}{c|c|c|c}
		\toprule
		
		  & JRDB-PAR   & NBA    & Volleyball  \\  \midrule
		Random Crops   &  40.7 & \textbf{83.3}   & 88.5 \\
		G.T. BB's   &  \textbf{44.9} &  -  & \textbf{93.1} \\ \bottomrule
	\end{tabular}}
 
    \label{tbl:ablation_crops}

\end{center}

\end{table}

\subsection{Qualitative Results}

\begin{figure}[ht]
    \centering
    \includegraphics[width=0.47\textwidth]{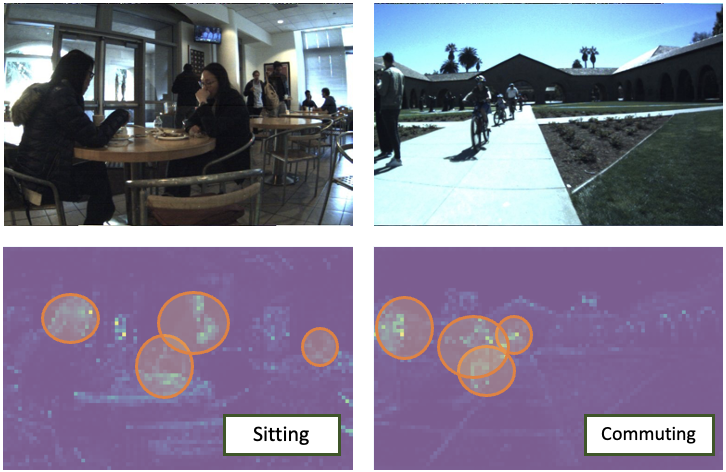}
    \caption{Visualization of the attention locations on the JRDB-PAR dataset. We show the locations of the top five attention weights from the transformer heads.}
    \label{fig:vis1}
\end{figure}

\begin{figure}[ht]
\begin{center}
\includegraphics[width=0.9\linewidth]{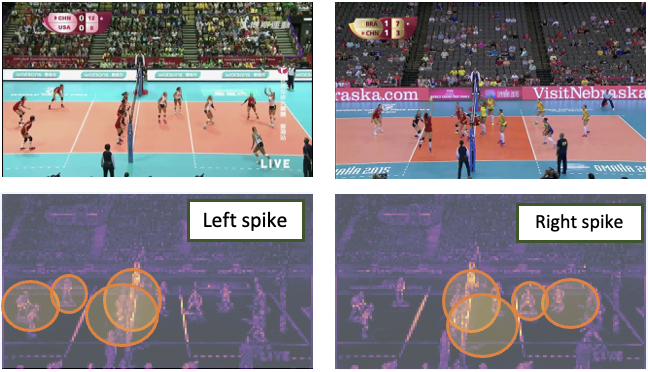}
\end{center}
\caption{Visualization of the attention locations on the Volleyball dataset. We show the locations of the top four attention weights from the transformer heads.}
\label{fig:vis2}
\end{figure}

We conducted an analysis to understand how our method aggregates feature for various social group activities. We visualized the attention locations of the transformer encoder in Fig.~\ref{fig:vis1} and Fig.~\ref{fig:vis2} for JRDB-PAR and Volleyball datasets, showing locations with the top five and top four attention weights in the last layer of the encoder. The yellow circles represent the attention locations. The size of the yellow circles denotes whether the locations are in the high or low-resolution feature maps, giving a rough indication of the image areas affecting the generated features. Our findings reveal that features are generally aggregated from low-resolution feature maps when group members are situated in broader areas, and the opposite is true. These results indicate that the proposed framework can effectively aggregate features based on the distribution of group members, thereby contributing to improving the performance of social group activity recognition.

\section{Conclusion}
\label{sec:conclusion}
Our paper presents a new self-supervised video model named SoGAR, which is based on a video transformer architecture. The method entails generating multiple views of a video, which differ in terms of their spatial and temporal characteristics. To capture the motion characteristics and cross-view relationships between the clips, we define two sets of correspondence learning tasks. The self-supervised objective is to reconstruct one view from another in the latent space of both the teacher and student networks. Furthermore, our SoGAR model can capture long-term spatio-temporal dependencies and perform dynamic inference within a single framework. We evaluate SoGAR on three benchmark datasets for social group activity recognition and demonstrate its superior performance over existing state-of-the-art models.

\section*{Acknowledgment}
We would like to acknowledge Arkansas Biosciences Institute (ABI) Grant, and NSF Data Science, Data Analytics that are Robust and Trusted (DART) for their funding in supporting this research.

{\small
\bibliographystyle{unsrt}
\bibliography{bibliography}

\begin{thebibliography}{10}

\bibitem{wang2016temporal}
Limin Wang, Yuanjun Xiong, Zhe Wang, Yu~Qiao, Dahua Lin, Xiaoou Tang, and Luc Van~Gool.
\newblock Temporal segment networks: Towards good practices for deep action recognition.
\newblock In {\em European Conference on Computer Vision}, pages 20--36. Springer, 2016.

\bibitem{carreira2017quo}
Joao Carreira and Andrew Zisserman.
\newblock Quo vadis, action recognition? a new model and the kinetics dataset.
\newblock In {\em Proceedings of the IEEE conference on computer vision and pattern recognition}, pages 6299--6308, 2017.

\bibitem{wang2018non}
Xiaolong Wang, Ross Girshick, Abhinav Gupta, and Kaiming He.
\newblock Non-local neural networks.
\newblock In {\em Proceedings of the IEEE conference on computer vision and pattern recognition}, pages 7794--7803, 2018.

\bibitem{Ranasinghe_2022_CVPR}
Kanchana Ranasinghe, Muzammal Naseer, Salman Khan, Fahad~Shahbaz Khan, and Michael~S. Ryoo.
\newblock Self-supervised video transformer.
\newblock In {\em Proceedings of the IEEE/CVF Conference on Computer Vision and Pattern Recognition (CVPR)}, pages 2874--2884, June 2022.

\bibitem{ibrahim2016hierarchical}
Mostafa~S Ibrahim, Srikanth Muralidharan, Zhiwei Deng, Arash Vahdat, and Greg Mori.
\newblock A hierarchical deep temporal model for group activity recognition.
\newblock In {\em Proceedings of the IEEE conference on computer vision and pattern recognition}, pages 1971--1980, 2016.

\bibitem{wu2019learning}
Jianchao Wu, Limin Wang, Li~Wang, Jie Guo, and Gangshan Wu.
\newblock Learning actor relation graphs for group activity recognition.
\newblock In {\em Proceedings of the IEEE conference on computer vision and pattern recognition}, pages 9964--9974, 2019.

\bibitem{hu2020progressive}
Guyue Hu, Bo~Cui, Yuan He, and Shan Yu.
\newblock Progressive relation learning for group activity recognition.
\newblock In {\em Proceedings of the IEEE conference on computer vision and pattern recognition}, pages 980--989, 2020.

\bibitem{gavrilyuk2020actor}
Kirill Gavrilyuk, Ryan Sanford, Mehrsan Javan, and Cees~GM Snoek.
\newblock Actor-transformers for group activity recognition.
\newblock In {\em Proceedings of the IEEE conference on computer vision and pattern recognition}, pages 839--848, 2020.

\bibitem{pramono2020empowering}
Rizard Renanda~Adhi Pramono, Yie~Tarng Chen, and Wen~Hsien Fang.
\newblock Empowering relational network by self-attention augmented conditional random fields for group activity recognition.
\newblock In {\em Computer Vision--ECCV 2020: 16th European Conference, Glasgow, UK, August 23--28, 2020, Proceedings, Part I 16}, pages 71--90. Springer, 2020.

\bibitem{ehsanpour2020joint}
Mahsa Ehsanpour, Alireza Abedin, Fatemeh Saleh, Javen Shi, Ian Reid, and Hamid Rezatofighi.
\newblock Joint learning of social groups, individuals action and sub-group activities in videos.
\newblock In {\em European Conference on Computer Vision}, pages 177--195. Springer, 2020.

\bibitem{yan2020higcin}
Rui Yan, Lingxi Xie, Jinhui Tang, Xiangbo Shu, and Qi~Tian.
\newblock Higcin: hierarchical graph-based cross inference network for group activity recognition.
\newblock {\em IEEE transactions on pattern analysis and machine intelligence}, 2020.

\bibitem{yuan2021learning}
Hangjie Yuan and Dong Ni.
\newblock Learning visual context for group activity recognition.
\newblock In {\em AAAI}, volume~35, pages 3261--3269, 2021.

\bibitem{li2021groupformer}
Shuaicheng Li, Qianggang Cao, Lingbo Liu, Kunlin Yang, Shinan Liu, Jun Hou, and Shuai Yi.
\newblock Groupformer: Group activity recognition with clustered spatial-temporal transformer.
\newblock {\em Proceedings of the IEEE international conference on computer vision}, 2021.

\bibitem{bagautdinov2017social}
Timur Bagautdinov, Alexandre Alahi, Fran{\c{c}}ois Fleuret, Pascal Fua, and Silvio Savarese.
\newblock Social scene understanding: End-to-end multi-person action localization and collective activity recognition.
\newblock In {\em Proceedings of the IEEE conference on computer vision and pattern recognition}, pages 4315--4324, 2017.

\bibitem{zhang2019fast}
Peizhen Zhang, Yongyi Tang, Jian-Fang Hu, and Wei-Shi Zheng.
\newblock Fast collective activity recognition under weak supervision.
\newblock {\em IEEE Transactions on Image Processing}, 29:29--43, 2019.

\bibitem{yan2020social}
Rui Yan, Lingxi Xie, Jinhui Tang, Xiangbo Shu, and Qi~Tian.
\newblock Social adaptive module for weakly-supervised group activity recognition.
\newblock In {\em European Conference on Computer Vision}, pages 208--224. Springer, 2020.

\bibitem{kim2022detector}
Dongkeun Kim, Jinsung Lee, Minsu Cho, and Suha Kwak.
\newblock Detector-free weakly supervised group activity recognition.
\newblock In {\em Proceedings of the IEEE/CVF Conference on Computer Vision and Pattern Recognition}, pages 20083--20093, 2022.

\bibitem{gberta_2021_ICML}
Gedas Bertasius, Heng Wang, and Lorenzo Torresani.
\newblock Is space-time attention all you need for video understanding?
\newblock In {\em ICML}, volume~2, page~4, 2021.

\bibitem{naseer2021intriguing}
Muzammal Naseer, Kanchana Ranasinghe, Salman Khan, Munawar Hayat, Fahad~Shahbaz Khan, and Ming-Hsuan Yang.
\newblock Intriguing properties of vision transformers.
\newblock {\em arXiv}, 2021.

\bibitem{han2022panoramic}
Ruize Han, Haomin Yan, Jiacheng Li, Songmiao Wang, Wei Feng, and Song Wang.
\newblock Panoramic human activity recognition.
\newblock In {\em Computer Vision--ECCV 2022: 17th European Conference, Tel Aviv, Israel, October 23--27, 2022, Proceedings, Part IV}, pages 244--261. Springer, 2022.

\bibitem{amer2012cost}
Mohamed~R Amer, Dan Xie, Mingtian Zhao, Sinisa Todorovic, and Song-Chun Zhu.
\newblock Cost-sensitive top-down/bottom-up inference for multiscale activity recognition.
\newblock In {\em European Conference on Computer Vision}, pages 187--200. Springer, 2012.

\bibitem{amer2013monte}
Mohamed~R Amer, Sinisa Todorovic, Alan Fern, and Song-Chun Zhu.
\newblock Monte carlo tree search for scheduling activity recognition.
\newblock In {\em Proceedings of the IEEE international conference on computer vision}, pages 1353--1360, 2013.

\bibitem{amer2014hirf}
Mohamed~Rabie Amer, Peng Lei, and Sinisa Todorovic.
\newblock Hirf: Hierarchical random field for collective activity recognition in videos.
\newblock In {\em European Conference on Computer Vision}, pages 572--585. Springer, 2014.

\bibitem{amer2015sum}
Mohamed~R Amer and Sinisa Todorovic.
\newblock Sum product networks for activity recognition.
\newblock {\em IEEE transactions on pattern analysis and machine intelligence}, 38(4):800--813, 2015.

\bibitem{lan2011discriminative}
Tian Lan, Yang Wang, Weilong Yang, Stephen~N Robinovitch, and Greg Mori.
\newblock Discriminative latent models for recognizing contextual group activities.
\newblock {\em IEEE transactions on pattern analysis and machine intelligence}, 34(8):1549--1562, 2011.

\bibitem{lan2012social}
Tian Lan, Leonid Sigal, and Greg Mori.
\newblock Social roles in hierarchical models for human activity recognition.
\newblock In {\em Proceedings of the IEEE conference on computer vision and pattern recognition}, pages 1354--1361. IEEE, 2012.

\bibitem{shu2015joint}
Tianmin Shu, Dan Xie, Brandon Rothrock, Sinisa Todorovic, and Song Chun~Zhu.
\newblock Joint inference of groups, events and human roles in aerial videos.
\newblock In {\em Proceedings of the IEEE conference on computer vision and pattern recognition}, pages 4576--4584, 2015.

\bibitem{wang2013bilinear}
Zhenhua Wang, Qinfeng Shi, Chunhua Shen, and Anton Van Den~Hengel.
\newblock Bilinear programming for human activity recognition with unknown mrf graphs.
\newblock In {\em Proceedings of the IEEE conference on computer vision and pattern recognition}, pages 1690--1697, 2013.

\bibitem{deng2016structure}
Zhiwei Deng, Arash Vahdat, Hexiang Hu, and Greg Mori.
\newblock Structure inference machines: Recurrent neural networks for analyzing relations in group activity recognition.
\newblock In {\em Proceedings of the IEEE conference on computer vision and pattern recognition}, pages 4772--4781, 2016.

\bibitem{ibrahim2018hierarchical}
Mostafa~S Ibrahim and Greg Mori.
\newblock Hierarchical relational networks for group activity recognition and retrieval.
\newblock In {\em European Conference on Computer Vision}, pages 721--736, 2018.

\bibitem{li2017sbgar}
Xin Li and Mooi Choo~Chuah.
\newblock Sbgar: Semantics based group activity recognition.
\newblock In {\em Proceedings of the IEEE international conference on computer vision}, pages 2876--2885, 2017.

\bibitem{qi2018stagnet}
Mengshi Qi, Jie Qin, Annan Li, Yunhong Wang, Jiebo Luo, and Luc Van~Gool.
\newblock stagnet: An attentive semantic rnn for group activity recognition.
\newblock In {\em European Conference on Computer Vision}, pages 101--117, 2018.

\bibitem{shu2019hierarchical}
Xiangbo Shu, Jinhui Tang, Guojun Qi, Wei Liu, and Jian Yang.
\newblock Hierarchical long short-term concurrent memory for human interaction recognition.
\newblock {\em IEEE transactions on pattern analysis and machine intelligence}, 2019.

\bibitem{wang2017recurrent}
Minsi Wang, Bingbing Ni, and Xiaokang Yang.
\newblock Recurrent modeling of interaction context for collective activity recognition.
\newblock In {\em Proceedings of the IEEE conference on computer vision and pattern recognition}, pages 3048--3056, 2017.

\bibitem{yan2018participation}
Rui Yan, Jinhui Tang, Xiangbo Shu, Zechao Li, and Qi~Tian.
\newblock Participation-contributed temporal dynamic model for group activity recognition.
\newblock In {\em Proceedings of the 26th ACM international conference on Multimedia}, pages 1292--1300, 2018.

\bibitem{yuan2021spatio}
Hangjie Yuan, Dong Ni, and Mang Wang.
\newblock Spatio-temporal dynamic inference network for group activity recognition.
\newblock In {\em Proceedings of the IEEE international conference on computer vision}, 2021.

\bibitem{chappa2024hatt}
Naga Venkata Sai~Raviteja Chappa, Pha Nguyen, Thi Hoang~Ngan Le, Page~Daniel Dobbs, and Khoa Luu.
\newblock Hatt-flow: Hierarchical attention-flow mechanism for group-activity scene graph generation in videos.
\newblock {\em Sensors}, 24(11):3372, 2024.

\bibitem{chappa2024ligar}
Naga Venkata Sai~Raviteja Chappa and Khoa Luu.
\newblock Ligar: Lidar-guided hierarchical transformer for multi-modal group activity recognition.
\newblock {\em arXiv preprint arXiv:2410.21108}, 2024.

\bibitem{chappa2024react}
Naga VS~Raviteja Chappa, Pha Nguyen, Page~Daniel Dobbs, and Khoa Luu.
\newblock React: recognize every action everywhere all at once.
\newblock {\em Machine Vision and Applications}, 35(4):102, 2024.

\bibitem{chappa2024flaash}
Naga~VS Chappa, Page~Daniel Dobbs, Bhiksha Raj, and Khoa Luu.
\newblock Flaash: Flow-attention adaptive semantic hierarchical fusion for multi-modal tobacco content analysis.
\newblock {\em arXiv preprint arXiv:2410.19896}, 2024.

\bibitem{truong2022otadapt}
Thanh-Dat Truong, Ravi Teja~Nvs Chappa, Xuan-Bac Nguyen, Ngan Le, Ashley~PG Dowling, and Khoa Luu.
\newblock Otadapt: Optimal transport-based approach for unsupervised domain adaptation.
\newblock In {\em 2022 26th international conference on pattern recognition (ICPR)}, pages 2850--2856. IEEE, 2022.

\bibitem{han2022dual}
Mingfei Han, David~Junhao Zhang, Yali Wang, Rui Yan, Lina Yao, Xiaojun Chang, and Yu~Qiao.
\newblock Dual-ai: Dual-path actor interaction learning for group activity recognition.
\newblock In {\em Proceedings of the IEEE/CVF Conference on Computer Vision and Pattern Recognition}, pages 2990--2999, 2022.

\bibitem{tamura2022hunting}
Masato Tamura, Rahul Vishwakarma, and Ravigopal Vennelakanti.
\newblock Hunting group clues with transformers for social group activity recognition.
\newblock In {\em Computer Vision--ECCV 2022: 17th European Conference, Tel Aviv, Israel, October 23--27, 2022, Proceedings, Part IV}, pages 19--35. Springer, 2022.

\bibitem{zhang2021multi}
Yanyi Zhang, Xinyu Li, and Ivan Marsic.
\newblock Multi-label activity recognition using activity-specific features and activity correlations.
\newblock In {\em Proceedings of the IEEE/CVF Conference on Computer Vision and Pattern Recognition}, pages 14625--14635, 2021.

\bibitem{vaswani2017attention}
Ashish Vaswani, Noam Shazeer, Niki Parmar, Jakob Uszkoreit, Llion Jones, Aidan~N Gomez, {\L}ukasz Kaiser, and Illia Polosukhin.
\newblock Attention is all you need.
\newblock In {\em NIPS}, pages 5998--6008, 2017.

\bibitem{dosovitskiy2020image}
Alexey Dosovitskiy, Lucas Beyer, Alexander Kolesnikov, Dirk Weissenborn, Xiaohua Zhai, Thomas Unterthiner, Mostafa Dehghani, Matthias Minderer, Georg Heigold, Sylvain Gelly, et~al.
\newblock An image is worth 16x16 words: Transformers for image recognition at scale.
\newblock {\em arXiv preprint arXiv:2010.11929}, 2020.

\bibitem{li2021ffa}
Mingjie Li, Wenjia Cai, Rui Liu, Yuetian Weng, Xiaoyun Zhao, Cong Wang, Xin Chen, Zhong Liu, Caineng Pan, Mengke Li, et~al.
\newblock Ffa-ir: Towards an explainable and reliable medical report generation benchmark.
\newblock In {\em Thirty-fifth Conference on Neural Information Processing Systems Datasets and Benchmarks Track (Round 2)}, 2021.

\bibitem{yuan2021tokens}
Li~Yuan, Yunpeng Chen, Tao Wang, Weihao Yu, Yujun Shi, Zihang Jiang, Francis~EH Tay, Jiashi Feng, and Shuicheng Yan.
\newblock Tokens-to-token vit: Training vision transformers from scratch on imagenet.
\newblock {\em arXiv preprint arXiv:2101.11986}, 2021.

\bibitem{liu2021swin}
Ze~Liu, Yutong Lin, Yue Cao, Han Hu, Yixuan Wei, Zheng Zhang, Stephen Lin, and Baining Guo.
\newblock Swin transformer: Hierarchical vision transformer using shifted windows.
\newblock In {\em Proceedings of the IEEE international conference on computer vision}, 2021.

\bibitem{wang2021pyramid}
Wenhai Wang, Enze Xie, Xiang Li, Deng-Ping Fan, Kaitao Song, Ding Liang, Tong Lu, Ping Luo, and Ling Shao.
\newblock Pyramid vision transformer: A versatile backbone for dense prediction without convolutions.
\newblock {\em arXiv preprint arXiv:2102.12122}, 2021.

\bibitem{han2020mining}
Mingfei Han, Yali Wang, Xiaojun Chang, and Yu~Qiao.
\newblock Mining inter-video proposal relations for video object detection.
\newblock In {\em European conference on computer vision}, pages 431--446. Springer, 2020.

\bibitem{arnab2021vivit}
Anurag Arnab, Mostafa Dehghani, Georg Heigold, Chen Sun, Mario Lu{\v{c}}i{\'c}, and Cordelia Schmid.
\newblock Vivit: A video vision transformer.
\newblock {\em arXiv preprint arXiv:2103.15691}, 2021.

\bibitem{li2022uniformer}
Kunchang Li, Yali Wang, Junhao Zhang, Peng Gao, Guanglu Song, Yu~Liu, Hongsheng Li, and Yu~Qiao.
\newblock Uniformer: Unifying convolution and self-attention for visual recognition, 2022.

\bibitem{bertasius2021space}
Gedas Bertasius, Heng Wang, and Lorenzo Torresani.
\newblock Is space-time attention all you need for video understanding?
\newblock In {\em ICML}, volume~2, page~4, 2021.

\bibitem{fan2021multiscale}
Haoqi Fan, Bo~Xiong, Karttikeya Mangalam, Yanghao Li, Zhicheng Yan, Jitendra Malik, and Christoph Feichtenhofer.
\newblock Multiscale vision transformers.
\newblock {\em arXiv preprint arXiv:2104.11227}, 2021.

\bibitem{patrick2021keeping}
Mandela Patrick, Dylan Campbell, Yuki~M Asano, Ishan Misra~Florian Metze, Christoph Feichtenhofer, Andrea Vedaldi, Jo~Henriques, et~al.
\newblock Keeping your eye on the ball: Trajectory attention in video transformers.
\newblock In {\em NeurIPS}, 2021.

\bibitem{feichtenhofer2019slowfast}
Christoph Feichtenhofer, Haoqi Fan, Jitendra Malik, and Kaiming He.
\newblock Slowfast networks for video recognition.
\newblock In {\em Proceedings of the IEEE international conference on computer vision}, pages 6202--6211, 2019.

\bibitem{kahatapitiya2021coarse}
Kumara Kahatapitiya and Michael~S Ryoo.
\newblock Coarse-fine networks for temporal activity detection in videos.
\newblock In {\em Proceedings of the IEEE conference on computer vision and pattern recognition}, 2021.

\bibitem{caron2021emerging}
Mathilde Caron, Hugo Touvron, Ishan Misra, Herv\'e J\'egou, Julien Mairal, Piotr Bojanowski, and Armand Joulin.
\newblock Emerging properties in self-supervised vision transformers.
\newblock In {\em Proceedings of the IEEE international conference on computer vision}, 2021.

\bibitem{ehsanpour2022jrdb}
Mahsa Ehsanpour, Fatemeh Saleh, Silvio Savarese, Ian Reid, and Hamid Rezatofighi.
\newblock Jrdb-act: A large-scale dataset for spatio-temporal action, social group and activity detection.
\newblock In {\em Proceedings of the IEEE/CVF Conference on Computer Vision and Pattern Recognition}, pages 20983--20992, 2022.

\bibitem{choi2011learning}
Wongun Choi, Khuram Shahid, and Silvio Savarese.
\newblock Learning context for collective activity recognition.
\newblock In {\em Proceedings of the IEEE conference on computer vision and pattern recognition}, pages 3273--3280. IEEE, 2011.

\bibitem{devlin2018bert}
Jacob Devlin, Ming-Wei Chang, Kenton Lee, and Kristina Toutanova.
\newblock Bert: Pre-training of deep bidirectional transformers for language understanding.
\newblock {\em arXiv preprint arXiv:1810.04805}, 2018.

\bibitem{grill2020bootstrap}
Jean-Bastien Grill, Florian Strub, Florent Altch{\'e}, Corentin Tallec, Pierre~H Richemond, Elena Buchatskaya, Carl Doersch, Bernardo~Avila Pires, Zhaohan~Daniel Guo, Mohammad~Gheshlaghi Azar, et~al.
\newblock Bootstrap your own latent: A new approach to self-supervised learning.
\newblock In {\em Advances in neural information processing systems}, 2020.

\bibitem{imagenet}
Olga Russakovsky, Jia Deng, Hao Su, Jonathan Krause, Sanjeev Satheesh, Sean Ma, Zhiheng Huang, Andrej Karpathy, Aditya Khosla, Michael Bernstein, Alexander~C. Berg, and Li~Fei-Fei.
\newblock Imagenet large scale visual recognition challenge.
\newblock {\em Int. J. Comput. Vis.}, 2015.

\bibitem{kingma15adam}
Diederik~P. Kingma and Jimmy Ba.
\newblock Adam: A method for stochastic optimization.
\newblock In {\em Int. Conf. Learn. Represent.}, 2015.

\bibitem{Steiner2021HowTT}
Andreas Steiner, Alexander Kolesnikov, Xiaohua Zhai, Ross Wightman, Jakob Uszkoreit, and Lucas Beyer.
\newblock How to train your vit? data, augmentation, and regularization in vision transformers.
\newblock {\em arXiv}, 2021.

\bibitem{chen2021mocov3}
Xinlei Chen*, Saining Xie*, and Kaiming He.
\newblock An empirical study of training self-supervised vision transformers.
\newblock {\em arXiv}, 2021.

\bibitem{lin2019tsm}
Ji~Lin, Chuang Gan, and Song Han.
\newblock Tsm: Temporal shift module for efficient video understanding.
\newblock In {\em Proceedings of the IEEE/CVF International Conference on Computer Vision}, pages 7083--7093, 2019.

\bibitem{liu2021video}
Ze~Liu, Jia Ning, Yue Cao, Yixuan Wei, Zheng Zhang, Stephen Lin, and Han Hu.
\newblock Video swin transformer.
\newblock {\em arXiv}, 2021.

\bibitem{azar2019convolutional}
Sina~Mokhtarzadeh Azar, Mina~Ghadimi Atigh, Ahmad Nickabadi, and Alexandre Alahi.
\newblock Convolutional relational machine for group activity recognition.
\newblock In {\em Proceedings of the IEEE conference on computer vision and pattern recognition}, pages 7892--7901, 2019.

\bibitem{chappa2023spartan}
Naga~VS Chappa, Pha Nguyen, Alexander~H Nelson, Han-Seok Seo, Xin Li, Page~Daniel Dobbs, and Khoa Luu.
\newblock Spartan: Self-supervised spatiotemporal transformers approach to group activity recognition.
\newblock In {\em Proceedings of the IEEE/CVF Conference on Computer Vision and Pattern Recognition}, pages 5157--5167, 2023.

\end{thebibliography}
}

\begin{IEEEbiography}[{\includegraphics[width=1in,height=1.25in,clip,keepaspectratio]{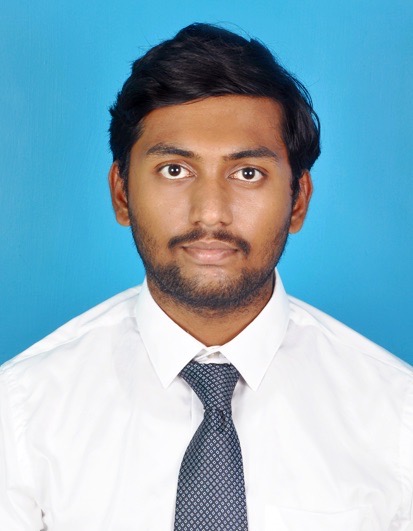}}]{Naga VS Raviteja Chappa} is currently a Ph.D. candidate in the Department of Computer Science and Computer Engineering at the University of Arkansas. He received his B.Tech degree in Electronics and Communication Engineering from MVGR College of Engineering (Affiliated to JNTU-K), India in 2018. He received his Master's degree with Computer Engineering major from Electrical and Computer Engineering at Purdue University, Indianapolis. Chappa's research interests focus on Vision-Language Models, Group Behavior Analysis \& their activity analysis, Video Understanding and Self-Supervised learning based methods. He also serves as a reviewer for CVPR, NeurIPS, ICLR, ICCV, ECCV, ICML, WACV, IEEE Access, MTAP Journal and Pattern Recognition.
\end{IEEEbiography}

\begin{IEEEbiography}[{\includegraphics[width=1in,height=1.25in,clip,keepaspectratio]{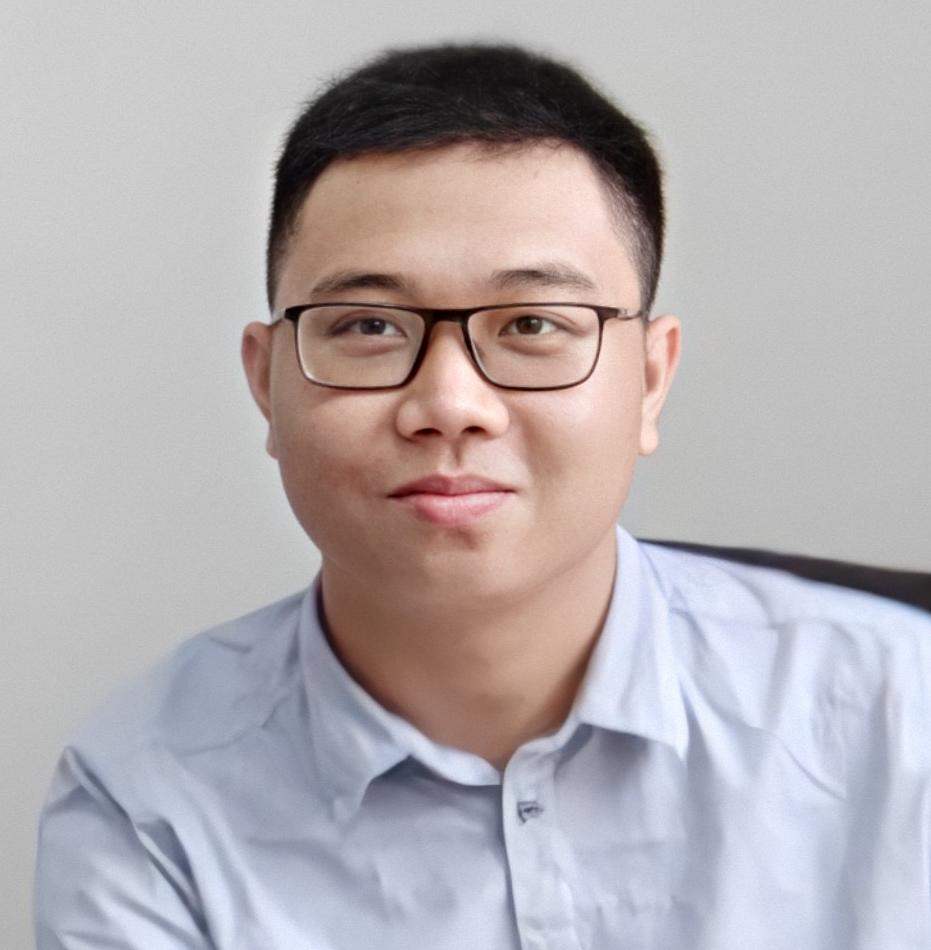}}]{Pha Nguyen} is currently a Ph.D student in Computer Science and a Research Assistant at Computer Vision and Image Understanding (CVIU) Lab, University of Arkansas, USA. In 2020, he obtained his B.Sc degree from Vietnam National University Ho Chi Minh City - University of Science in Computer Science. His particular research interests are focused on deep learning, machine learning, and their applications in computer vision, primarily multiple object tracking. He also serves as a reviewer for CVPR, ICCV, ECCV and IEEE Access.
\end{IEEEbiography}

\begin{IEEEbiography}[{\includegraphics[width=1in,height=1.25in,clip,keepaspectratio]{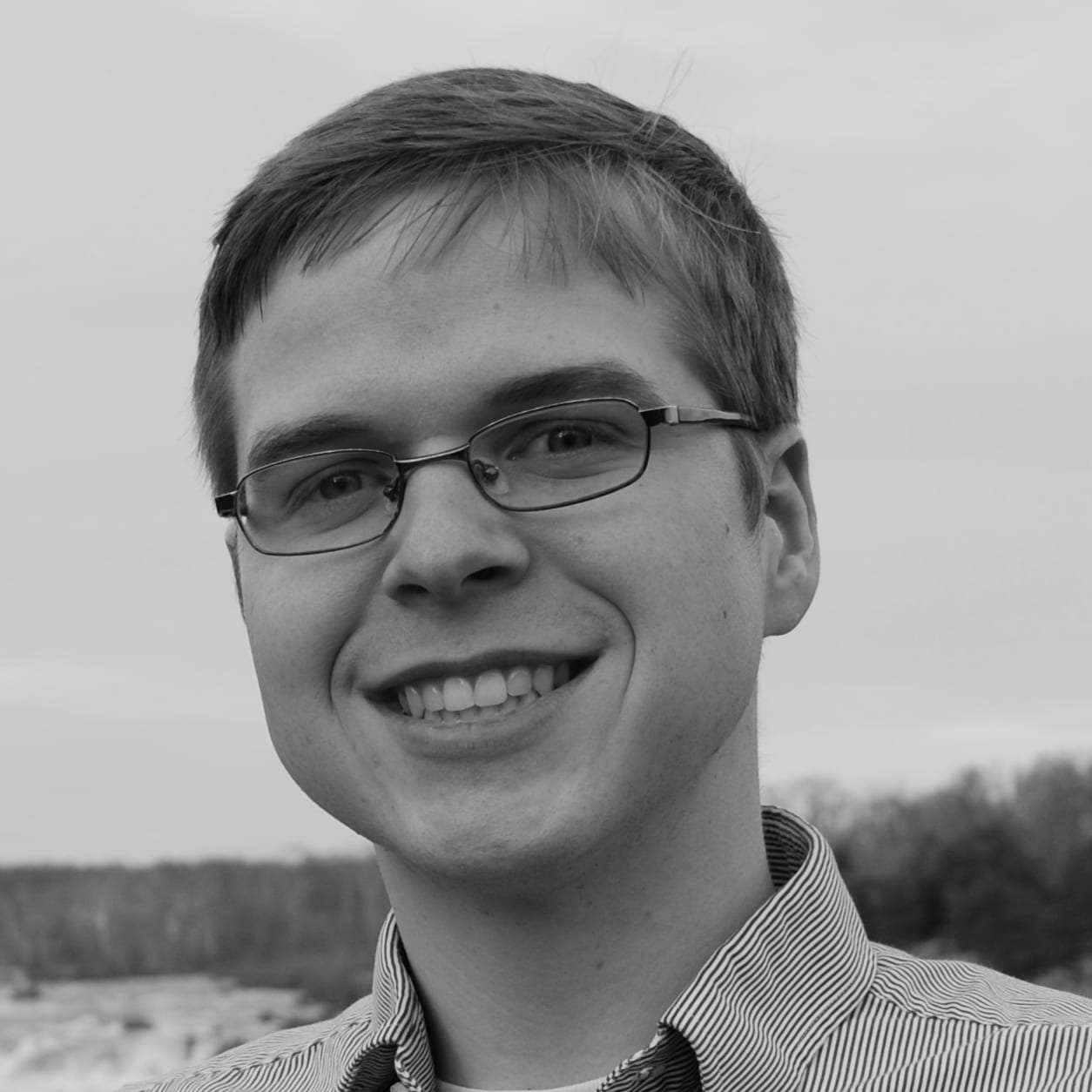}}]{Alexander H Nelson} is an Assistant Professor in the department of Computer Science and Computer Engineering at the University of Arkansas. His primary research interests include wearable and ubiquitous computing. He received his Ph.D. from the University of Maryland, Baltimore County (UMBC) in 2017.  At the University of Arkansas, Alexander leads the ÆSIR (Applied Embedded Systems and IoT Research) Laboratory.
\end{IEEEbiography}

\begin{IEEEbiography}[{\includegraphics[width=1in,height=1.25in,clip,keepaspectratio]{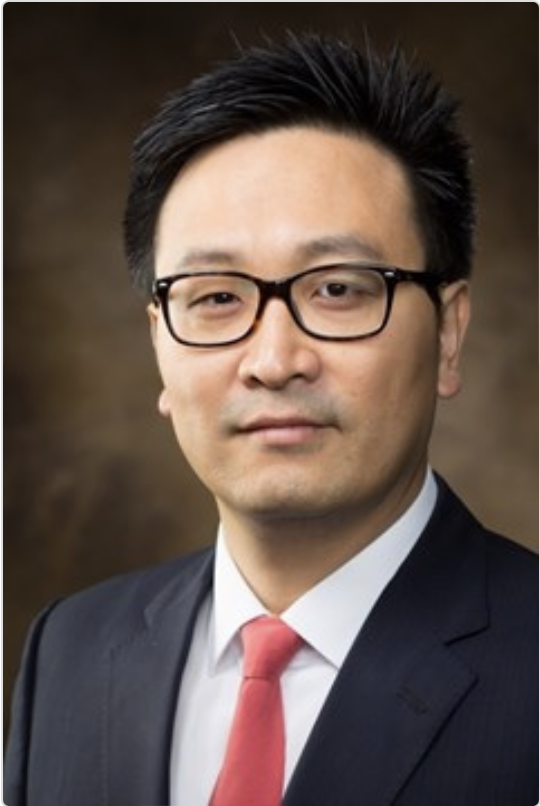}}]{ Han-Seok Seo} is an Associate Professor and Director of the University of Arkansas Sensory Science Center in the Department of Food Science at the University of Arkansas, Fayetteville. Dr. Seo is a creative sensory scientist who combines multidisciplinary backgrounds and skills in order to contribute to improved quality of life through healthy and happy eating behavior. His research interests include identifying mechanisms of multisensory interaction and integration with a focus on chemosensory cues, developing methods to improve eating quality, creating novel methodology of sensory evaluation, and investigating impacts of sensory disorders on eating quality. He holds two doctoral degrees, a Ph.D. in Food and Nutrition and a Doctor of Medical Science in Otorhinolaryngology from Seoul National University (Seoul, Korea) and the Technical University of Dresden (Dresden, Germany), respectively. Dr. Seo has published more than 120 articles in peer-reviewed journals, and he serves as an editorial board member of multiple journals including the Journal of Sensory Studies, Food Quality and Preference, Foods, Journal of Culinary Science and Technology, Journal of Food Science, and Korean Journal of Food and Cookery Science. He also serves as an Associate Editor and a Section Editor of the Food Research International and Current Opinion in Food Science, respectively.
\end{IEEEbiography}

\begin{IEEEbiography}[{\includegraphics[width=1in,height=1.25in,clip,keepaspectratio]{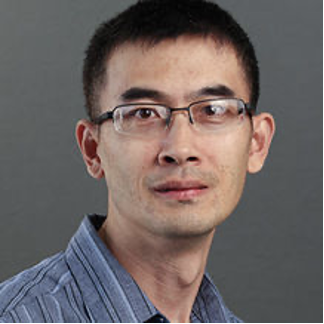}}]{Xin Li}
 received the B.S. degree with highest honors in electronic engineering and information science from University of Science and Technology of China, Hefei, in 1996, and the Ph.D. degree in electrical engineering from Princeton University, Princeton, NJ, in 2000. He was a Member of
Technical Staff with Sharp Laboratories of America, Camas, WA from Aug. 2000 to Dec. 2002. Since Jan. 2003, he has been a faculty member in Lane Department of Computer Science and Electrical Engineering. Dr. Li was elected a Fellow of IEEE in 2017 for his contributions to image interpolation, restoration and compression.
\end{IEEEbiography}

\begin{IEEEbiography}[{\includegraphics[width=1in,height=1.25in,clip,keepaspectratio]{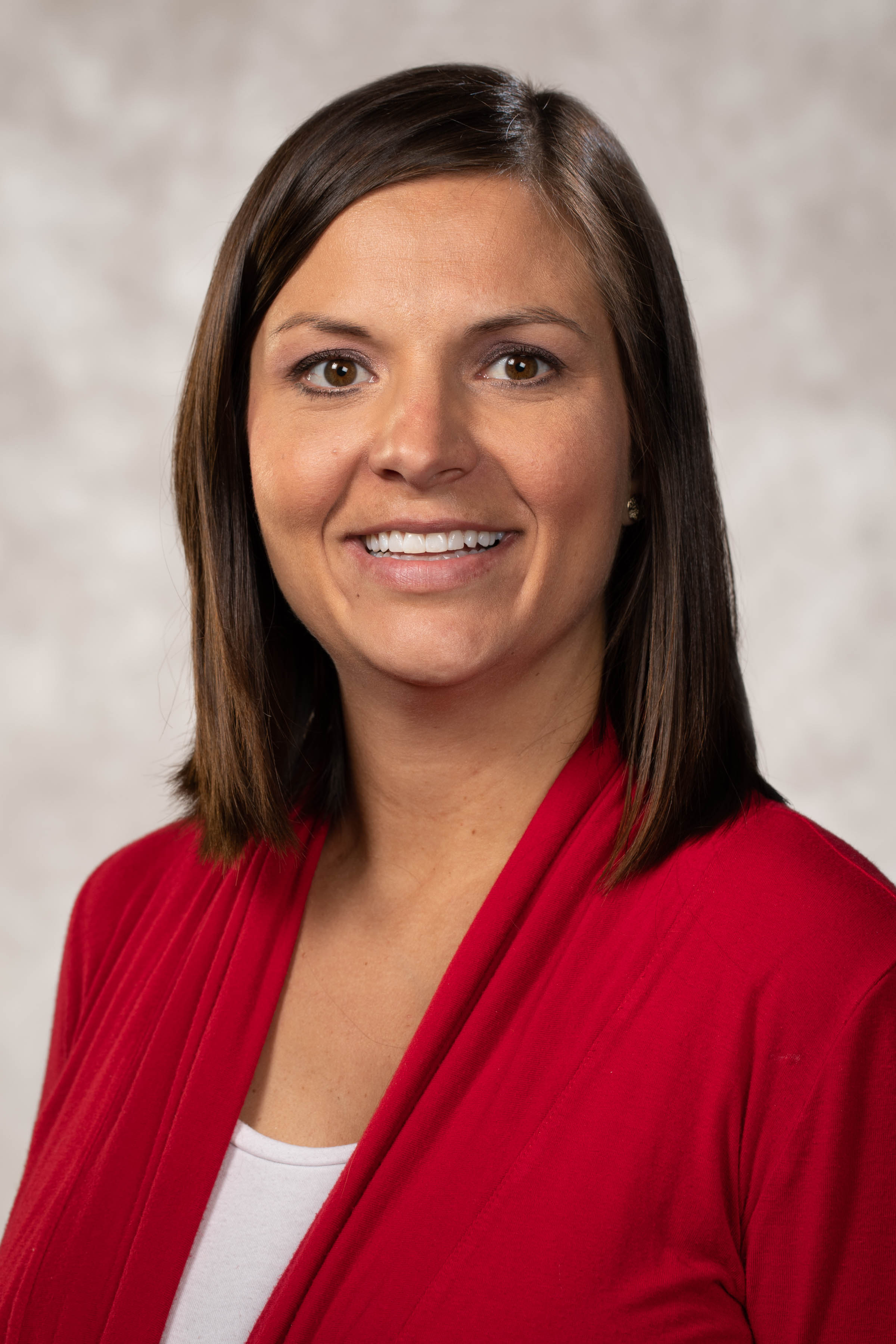}}]{Page Daniel Dobbs} earned a MS and a PhD in Community Health Promotion from the University of Arkansas’ College of Education and Health Professions. She joined the University of Arkansas faculty as an Assistant Professor of Public Health in the Department of Health, Human Performance and Recreation, in 2020. Dr. Dobbs' research focuses on social factors that influence perceptions and behaviors among vulnerable populations. She has published quantitative, qualitative, and mixed-methods research that examined the use of cigarettes and electronic cigarettes (e-cigarettes) among youth, young adults, minority populations, and pregnant women. Dr. Dobbs recently received a career award (Loopholes, Enforcement Challenges and Tobacco Industry Interference with Tobacco Control Policies) from the National Institute of Health,  entitled, “Loopholes, Enforcement Challenges and Tobacco Industry Interference with Tobacco Control Policies.”
\end{IEEEbiography}

\begin{IEEEbiography}[{\includegraphics[width=1in,height=1.25in,clip,keepaspectratio]{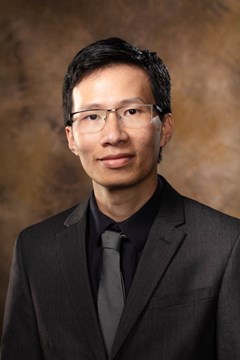}}]{Khoa Luu} Dr. Luu is currently an Assistant Professor and the Director of Computer Vision and Image Understanding (CVIU) Lab in Department of Computer Science \& Computer Engineering at University of Arkansas, Fayetteville. 
He was the Research Project Director in Cylab Biometrics Center at Carnegie Mellon University (CMU), USA. He has received four patents and two best paper awards and coauthored 100+ papers in conferences and journals. He was a vice-chair of Montreal Chapter IEEE SMCS in Canada from September 2009 to March 2011.
He is teaching Computer Vision, Image Processing and Introduction to Artificial Intelligence courses in 
CSCE Department at University of Arkansas, Fayetteville.
His research interests focus on various topics, including Biometrics, Image Processing, Computer Vision, Machine Learning, Deep Learning, Multifactor Analysis and Compressed Sensing. 
He is an co-organizer and a chair of CVPR Precognition Workshop in 2019, 2020, 2021; MICCAI Workshop in 2019, 2020 and ICCV Workshop in 2021. He is a PC member of AAAI, ICPRAI in 2020 and 2021.
He is currently a reviewer for several top-tier conferences and journals, such as CVPR, ICCV, ECCV, NeurIPS, ICLR, FG, BTAS, IEEE-TPAMI, IEEE-TIP, Journal of Pattern Recognition, Journal of Image and Vision Computing, Journal of Signal Processing, Journal of Intelligence Review, IEEE Access Trans., etc.
\end{IEEEbiography}

\EOD

\end{document}